\UseRawInputEncoding

\documentclass[journal]{IEEEtran}

\ifCLASSINFOpdf

\else
 
\fi

\usepackage{array}
\usepackage{amsmath}
\usepackage{graphicx}
\usepackage{upgreek}
\usepackage{multirow}
\usepackage{color}
\usepackage{pifont}
\usepackage{url}
\usepackage{bm}

\begin{document}

\title{ 3D-ANAS: 3D Asymmetric Neural Architecture Search for Fast Hyperspectral Image Classification}

\author{Haokui~Zhang,~Chengrong~Gong,~Yunpeng~Bai,~Zongwen Bai,and~Ying~Li
\thanks{Manuscript received xx xx, xxxx; revised xx xx, xxxx. This work was supported by the National Natural Science Foundation of China（61871460）， the Shaanxi Provincial Key R\&D Program （2020KW-003）， the Ministry of Science and Technology Foundation funded project (2020AAA0106900)}
\thanks{H. Zhang, C. Gong, Y. Bai, Z. Bai and Y. Li are with the Shaanxi Provincial Key Lab of Speech and Image Information Processing, School of Computer Science, Northwestern Polytechnical University, Xi'an 710129, China (e-mail: hkzhang1991@mail.nwpu.edu.cn; gongchengrong@mail.nwpu.edu.cn, cloudbai@nwpu.edu.cn, ydbzw@yau.edu.cn, lybyp@nwpu.edu.cn, ). Part of the work was done when H. Zhang was visiting University of Adelaide. (Corresponding author: Ying Li.)}
}

%
%

\markboth{IEEE TRANS XXXX}%
{Shell \MakeLowercase{\textit{et al.}}: Bare Demo of IEEEtran.cls for IEEE Journals}
%



\maketitle

\begin{abstract}

Hyperspectral images involve abundant spectral and spatial information, playing an irreplaceable role in land-cover classification. Recently, based on deep learning technologies, an increasing number of HSI classification approaches have been proposed, which demonstrate promising performance. However, previous studies suffer from two major drawbacks: 1) the architecture of most deep learning models is manually designed, relies on specialized knowledge, and is relatively tedious. Moreover, in HSI classifications, datasets captured by different sensors have different physical properties. Correspondingly, different models need to be designed for different datasets, which further increases the workload of designing architectures; 2) the mainstream framework is a patch-to-pixel framework. The overlap regions of patches of adjacent pixels are calculated repeatedly, which increases computational cost and time cost. Besides, the classification accuracy is sensitive to the patch size, which is artificially set based on extensive investigation experiments. To overcome the issues mentioned above, we firstly propose a 3D asymmetric neural network search algorithm and leverage it to automatically search for efficient architectures for HSI classifications. By analysing the characteristics of HSIs, we specifically build a 3D asymmetric decomposition search space, where spectral and spatial information are processed with different decomposition convolutions. Furthermore, we propose a new fast classification framework, i,e., pixel-to-pixel classification framework, which has no repetitive operations and reduces the overall cost. Experiments on three public HSI datasets captured by different sensors demonstrate the networks designed by our 3D-ANAS achieve competitive performance compared to several state-of-the-art methods, while having a much faster inference speed. Code is available at: \url{ https://github.com/hkzhang91/3D-ANAS}. 
\end{abstract}

\begin{IEEEkeywords}
Hyperspectral image classification, asymmetric network search, asymmetric search space, patch-to-pixel framework, pixel-to-pixel framework, inference speed.
\end{IEEEkeywords}

%
\IEEEpeerreviewmaketitle

\section{Introduction}

\IEEEPARstart{H}{yperspectral} images (HSIs) are captured via hyperpsectral remote sensors, which collect spatial information and spectral information in hundreds of spectral bands, covering a wide range of wavelengths at the same time. As the products of hyperspectral remote sensors, HSIs have rich spectral and spatial information, which are more effective in distinguishing different land-covers compared with other common remote images. Therefore, HSIs are widely employed in various applications due to their advanced features. Identifying the category of each pixel in HSI is a crucial technique for these applications. However, the rich spectral information in HSI improves Distinguishability on the one hand and it increases the dimensionality of samples one the other hand, which may affect the classification performance \cite{zhu2018multiple, brown2010hydrothermal, brown2008marte}. 

To solve this issue, various feature extraction approaches have been introduced into HSI classification. In the early stage, most of feature extraction approaches used in classification frameworks are handcrafted feature extractors. During this stage, two main frameworks are proposed one after another. The older one is the spectral information based classification, where only spectral information is exploited. Methods that use only spectral information fail in the the confronting two situations: 1) different objects exhibit similar spectral features; 2) the same objects located in different locations emerge with different spectral features. As a result, these methods generally show a relatively poor performance. Latter, both spectral and spatial information are employed to improve classification accuracy. Spectral and spatial information based HSI classification methods have two main types. The first one uses the spectral and spatial information separately. For instance, the spatial dependence is extracted in advance through spatial filters, such as morphological profiles and attribute profiles. Then, spatial dependence information is combined with spectral information. Alternatively, the spatial information can be exploited to refine the classification results obtained by using spectral information only via a regularization process, such as Markov Random Field (MRF) and graph cut. The other one is to combine spectral and spatial information to generate spectral-spatial features, which are then used to predict classification results. For example, several 3D filters (eg, 3D wavelet filters, 3D Gabor filters [24], and 3D scattering wavelet filers) have been applied on HSIs to extract spectral-spatial features. 

Recently, inspired by the great performance of deep learning techniques, various deep learning models, such as Stacked Autoencoder (SAE), Deep Belief Network (DBN), Recurrent Neural Network (RNN) and Convolutional Neural Network (CNN) have been introduced into HSI classification, leading to significant achievements~\cite{chen2014deep,chen2015spectral,zhang2017spectral,li2017spectral}. Learning from the experience accumulated in early stage in HSI classification, most of deep learning based HSI classification methods combine spectral and spatial information to accomplish the classification task. More specifically, one of the most significant advantages of deep learning is the ability to extract efficient deep features from raw images directly. This advantage is fully exploited in deep learning based classification methods, where deep learning models are applied to extract deep spectral-spatial features from raw HSI data. For instance, Chen \emph{et al}. applied unsupervised deep feature learning model SAE to extract spectral and spatial feature separately, and then combined the extracted features to generate spectral-spatial features for classification \cite{chen2014deep}. Later, Chen \emph{et al}. replaced SAE with DBN and proposed DBN based spectral-spatial feature extraction framework \cite{chen2015spectral}. Besides these unsupervised deep learning models, supervised CNN models have been exploited to extract spectral-spatial features. In~\cite{zhang2017spectral, zhong2018spectral,shu2018hyperspectral}, 2DCNN is developed to extract spectral-spatial features from compressed HSI data. In~\cite{2019Hyperspectral}, Zhang \emph{et al.} proposed a deep 3D light weight convolutional network which consists of tens of 3D convolution layers. Compared with SAE and DBN, CNN based HSI claffification methods yield better performance and have become the mainstream of deep learning based approaches for HSI classificaion in recent four years. 

Most classification approaches are based on manually designed deep learning models. However,investigating state-of-the-art neural network is not trivial and it requires substantial efforts, especially in HSI classification, different HSIs are collected by different sensors with different spatial resolutions, different number of spectral bands, different spectral ranges, etc. Correspondingly, different HSIs require different models to process, which further increases the workload of designing architectures. In addition, previous work has typically employed a patch-to-pixel framework, where a patch is cropped from the original HSI and all the information of this patch is fed to the trained model to identify a single pixel located at the center of the cropped patch. To obtain the classification result of an entire HSI with spatial resolution of $M\times N$, a sliding window strategy is adopted to crop $M\times N$ patches. Then, these cropped patches are fed into trained model for classification. In the patch-to-pixel framework, the overlap regions of different patches are calculated repeatedly, which increases computational cost and time cost. Besides, the classification accuracy is sensitive to the size of cropped patches. larger patches imply richer contextual information (spatial information). However, larger patches also represent a lower proportion of information in pixels for the central pixel in the patch. The patch size which is too small or too large is not conducive to improve classification accuracy. As a result, it is difficult to discover a proper patch size.  

In this paper, we attempt to address aforementioned issues from two aspects. Firstly, we propose a three-dimensional asymmetric neural architecture search (3D-ANAS) algorithm and leverage it to eliminate the tedious and heuristic procedure of manually design of network architectures from deep learning based pipeline. Recent three years, a growing interest is witnessed in developing algorithmic solutions to automate the process of architecture design. Architectures found by NAS algorithms have achieved highly comparable performance in both high-level vision tasks~\cite{liu2019auto, liu2018darts},  and low-level vision tasks~\cite{zhang2020memory, liu2019deep, chu2019fast}. Very recently, chen \emph{et al}. proposed automatic 1D Auto-CNN and 3D Auto-CNN for HSI classification and achieved encouraging performance~\cite{chen2019automatic}. Compared with manually designed architectures, the architectures of 1D Auto-CNN and 3D Auto-CNN outperformed with fewer parameters. However, in~\cite{chen2019automatic}, like most of other NAS works, only topological structures are designed via NAS algorithms. The widths (the numbers of channels of convolution layers) of different layers of the network  are manually set. In addition, the search space built in~\cite{chen2019automatic} still has room for improvement. In this work, we build a hierarchical search space including inner search space and outer search space, which are in charge of designing cell architectures and deciding cell widths, respectively. For inner search space, we build an 3D asymmetric decomposition search space. For outer search space, we employ cell sharing strategy to save memory resources. Secondly, we adopt pixel-to-pixel classification framework to improve the inference efficiency. Within the pixel-to-pixel classification framework, the trained model can take patches with arbitrary size as input and identify all pixels of the input patch at one time, which significantly enhanced the inference speed.

In summary, the work of this paper focuses on automatically searching for 3D asymmetric networks for fast HSI classification. Our main contributions are summarized as follows.

\begin{enumerate}[]

\item Based on the gradient search strategy, we propose a 3D asymmetric neural architecture search algorithm, termed 3D-ANAS. The proposed 3D-ANAS has a hierarchical search space which includes inner search space and outer search space. 3D-ANAS is capable of searching for both topological structure and network widths. By analyzing the characteristics of HSIs, we establish a 3D asymmetric decomposition search space as inner search space. To save memory, we employ cell sharing strategy in outer search space. 

\item In order to overcome the duplicate operation problem in the patch-to-pixel classification framework employed in previous works, we propose a pixel-to-pixel classification framework, which increases the inference speed to 10$\times$ of 3D Auto-CNN (which represents the state-of-the-art NAS based HSI classification method).

\item Aiming to the proposed pixel-to-pixel classification framework, we propose two augmentation inference strategies, which further improve the classification performance. Experimental results on three HSI datasets captured by different sensors after 2001 demonstrated that architectures found by our proposed 3D-ANAS outperform other manually designed 3D CNN based HSI classification methods and NAS based HSI classification methods in classification accuracy, while having much faster inference speed. 

\end{enumerate}

The rest of this paper is organized as follows. Section II presents the related work, consisting of deep learning based HSI classification methods and neural architecture search algorithms. Section III elaborates on our proposed method, including inner topological architecture search, outer widths search, pixel-to-pixel classification framework and training strategy. In Section IV, we first describe datasets and experimental setups; then we compare and analyze the experimental results; finally, we conclude this article and point out our future work in Section V.

\section{Related Work}

\subsection{DL for HSI classification}

Broadly speaking, there are four basic types of deep learning models, including Stacked Auto-encoder (SAE), Deep Belief Network (DBN), Convolutional Neural Network (CNN) and Recurrent Neural Network (RNN). Since 2013, when Lin \emph{et al}. proposed to use SAE to extract spectral-spatial features and achieved impressive classification performance, an increasing number of deep learning based HSI classification methods have been investigated. Until now, all four types of deep learning models and corresponding derivative models have been introduced to HSI classification. 

Earlier work exploited SAE and DBN to extract deep features for HSI classification. To our knowledge, the earliest attempt can be found in \cite{lin2013spectral}, where Lin and Chen \emph{et al}. proposed to use SAE to extract spectral-spatial features from preprocessed HSIs. The preprocessing operation is spectral compression, where principal component analysis (PCA) \cite{PCA2011} is employed to reduce the data dimensionality by compressing the raw HSI containing hundreds of spectral bands into a compressed HSI which consists of tens of spectral bands. Based on this work~\cite{lin2013spectral}, Chen \emph{et al}. proposed new SAE based HSI classification framework, where both compressed HSI and raw HSI can be utilized \cite{chen2014deep}. Specifically, the features extracted via the framework proposed in \cite{lin2013spectral} are regarded as spatial features, while the 1D features extracted from raw HSI with SAE are regarded as spectral features. Once spatial features and spectral features are extracted, they are combined to generate spectral-spatial features. Chen \emph{et al}. replaced SAE with DBN in the framework proposed in \cite{chen2014deep} and proposed DBN based spectral-spatial HSI classification method\cite{chen2015spectral}. In \cite{ma2016spectral}, Ma \emph{et al}. exploited SAE to learn the effective features, where a relative distance prior is added into the subsequent fine-tuning process. In the above four methods, spectral information does not require any pre-processing, but spatial information must be flattened to 1D vector, as SAE and DBN only accept 1D input. However, the flattened features do not retain the same spatial information that the original image may contain. 

In recent five years, various CNN models have been employed to classify HSIs. CNN based HSI classification methods were proposed later compared with SAE and DBN. While, statistically, the number of methods using CNNs for HSI classification has grown fastest and the performance of CNNs is generally superior. After 2016, CNN based methods have become the mainstream of deep learning based HSI classification approaches. Generally speaking, CNNs contains three main types, including 1D-CNN, 2D-CNN and 3D-CNN. In the 1D-CNN based work, convolution layers implement convolution operation on the input samples along the spectral dimension. For instance, Hu \emph{et al}. built a 1D-CNN network containing one convolution layer, one polling layer and two fully connected layers for HSI classification. Zhang \emph{et al}. proposed to use a 1D-CNN to extract spectral-spatial features from $3\times 3$ pixels~\cite{2016Spectral}. Meanwhile, Mei \emph{et al}. exploited a similar 1D-CNN to classify HSIs in \cite{mei2016integrating}. In 2D-CNN based classification methods, 2D-CNN models and dimension reduction algorithms are always used jointly.  Specifically, HSIs are always compressed via dimension reduction algorithm, and then convolved with 2D kernels. For example, in \cite{makantasis2015deep}, R-PCA is used to condense the entire HSI, then 2D CNN is employed to extract spatial features. Later, a similar framework is proposed in \cite{yue2015spectral}, where the top three principal components were extracted from the raw HSI by using PCA, then fed into 2D-CNN to extract spatial features. With 3D-CNNs, robust spectral-spatial features can be extracted from raw HSIs directly, as both the model and the data have a three-dimensional structure. In fact, compared with 1D-CNN and 2D-CNN based methods, 3D-CNN based methods generally have higher classification accuracy. In \cite{chen2016deep} and \cite{li2017spectral}, 3D-CNNs were adopted to extract spectral-spatial features from HSI data without any preprocessing. In \cite{zhong2018spectral}, Zhong \emph{et al.} employed spectral and spatial residual blocks consecutively to learn spectral and spatial representations separately. Very recently, Zhang \emph{et al}. proposed a deep 3D light weight convolutional network and combined it with transfer learning to perform HSI classification in the condition of limited training samples \cite{2019Hyperspectral}. RNN is mainly designed to handle sequential data. Compared with SAE, DBN and CNN, approaches based on RNNs are relatively few. In \cite{wu2017convolutional}, Wu \emph{et al}. proposed a convolutional recurrent neural network (CRNN) for HSI classification, consisting of a few convolution layers followed by recurrent layers.

Broadly speaking, 3D-CNNs yield superior performance in HSI classification compared with SAE, DBN and RNN. However, all the 3D-CNN based HSI classification methods introduced above are still based on manually designed architectures.

\subsection{Neural architecture search}

Recently, NAS algorithms have attracted sidespread attention and outperformed manually designed architectures on various computer vision tasks, including image classification\cite{zoph2018learning}, semantic segmentation~\cite{liu2019auto}, object detection~\cite{NASFCOS}, image denoising~\cite{zhang2020memory, suganuma2018exploiting, liu2019deep}, image super-resolution~\cite{song2020efficient}, designing data augmentation strategy~\cite{2020AutoAugment}, designing activation function~\cite{Ramachandran2017Searching}, etc. Differing from conventional deep learning based image processing work, where architectures of networks are designed manually, NAS aim to design automated approaches to discover high-performance neural architectures. Until now, there are three mainstream search strategies, including evolutionary algorithms (EAs), reinforcement learning (RL), and gradient-based architecture search. RL based works, in \cite{zoph2018learning} and \cite{zhong2018practical}, RL techniques policy gradients and Q-learning are respectively employed to train a recurrent neural network that works as meta-controller to generate potential architectures. \cite{real2019regularized, liu2017hierarchical, song2020efficient} evolve a set of stochastic construction models by EA to obtain better architecture. EA and RL based methods are normaly inefficient in search,  often requiring a large amount of computations. To remedy this deficiency, multiple speed-up techniques are proposed. For instance, hyper-networks \cite{zhang2018graph}, network morphism  \cite{elsken2018efficient} and shared weights \cite{pham2018efficient}. Gradient based NAS methods were proposed in recent two years. Compared with previous two types, gradient based NAS methods have been developed for a shorter time. The earliest attempt can be found in \cite{liu2018darts}, where DARTS was proposed for the first time. Differing from EA and RL based methods which train a huge number of student networks, DARTS just train a single supernet in the search phase, enabling the searching stage to be quite efficient. Following DARTS, various gradient based NAS methods have been proposed for image classification~\cite{cai2019device}, semantic segmentation~\cite{liu2019auto}, image restoration~\cite{zhang2020memory, chu2019fast} and HSI classification~\cite{chen2019automatic}. 

Our proposed 3D-ANAS is most closely related to automatic 3-D Auto-CNN~\cite{chen2019automatic}. Both are gradient based NAS methods and are proposed for HSI classification. Automatic 3-D Auto-CNN build a 2D convolution search space which is similar with that of DARTS. In addition, similar to previous 2D CNN based HSI classification methods, automatic 3-D Auto-CNN has a preprocessing stage, in which raw HSI is condensed along the spectral dimension via a point wise convolution. Compared with automatic 3-D Auto-CNN, our proposed 3D-ANAS has four main differences: 1) 3D asymmetric decomposition inner search space. By analyzing the characteristics of HSI, we build a 3D asymmetric decomposition inner search space, where 3D asymmetric decomposition convolution operations are employed. Since both operations in the search space and HSI data are three-dimensional structures, no preprocessing operations are required in our framework; 2) search for different architectures for different layers, increasing flexibility ; 3) hierarchical search space. We build a hierarchical search space which can search both topological architecture and network width; 4) a pixel-to-pixel classification framework. To overcome the problem of repetitive operation, we propose a pixel-to-pixel classification framework, which significantly improves the inference speed.

\section{Proposed Method}

As indicated previously, our proposed 3D-ANAS has a hierarchical search space which consists of an inner search space and an outer search space. By analyzing the characteristics of HSIs, we build a 3D asymmetric decomposition search space as the inner search space, which is used to search for topological architectures. The outer search space contains three candidate paths regarding to different widths. The outer search space is employed to decide the widths for different layers. In addition, the classification framework used in this paper is a pixel-to-pixel classification framework. 

In this section, we first present our 3D asymmetric decomposition search space. Then we explain how to search for widths for different layers in the outer search space. Finally, we introduce the pixel-to-pixel classification framework and our training strategy.

\subsection{Inner Topological Architecture Search} 
\label{Sec: inner_search}

\begin{figure}[t]
\setlength{\abovecaptionskip}{0.cm}
\setlength{\belowcaptionskip}{-0.cm}
\centering
\includegraphics[width=3.4in]{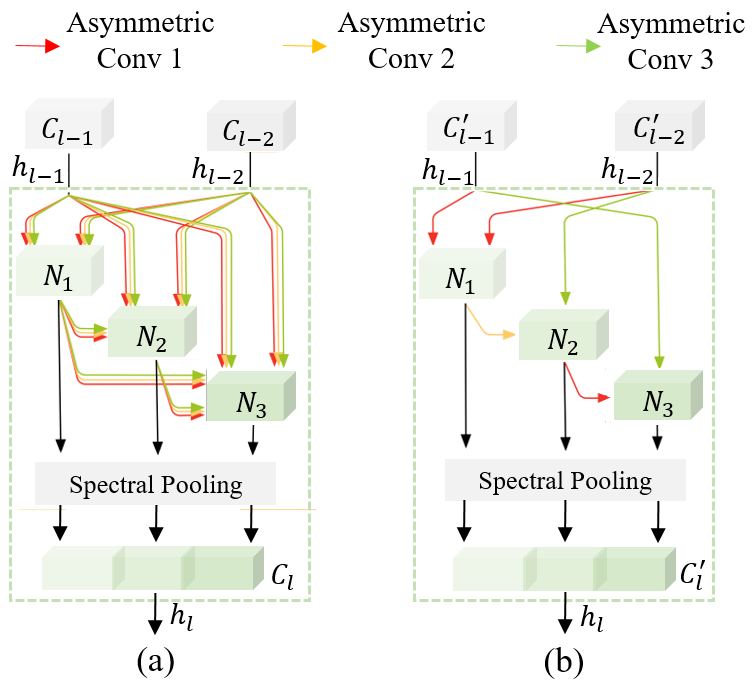}
\caption{The 3D-asym-d search space and topological architecture search. Left: 3D-asym-d search space. For each layer, we build a supercell that consists of all possible operation types, which is represented as arrows with different colors. Right: the result of topological architecture search, a compact cell, where each node only keeps the two top valuable inputs, each of which is feed to the selected operation. Here, we only show three nodes and three asymmetric convolution operations for clear exposition.}
\label{fig: inner_search}
\end{figure}

\vspace{3 pt}
\noindent {\bf 3D-asym-d Search Space}
Differing from previous works~\cite{zhang2020memory, liu2018darts, chen2019automatic}, which search for a single computation cell as the basic block and then build the whole network by stacking the found block with different widths, our proposed 3D-ANAS does not share architecture across different layers, it searches for $L$ different cells for the final network which contains $L$ layers and then builds the final network by stacking the $L$ cells in order. To be specific, for each layer, we build three supercells, which have different widths and each contains a sequence of $N$ nodes. Each node is connected to two inputs of the current supercell and nodes before it processes through different paths. In the supercell, each path has all candidate operations, each of which corresponds to a learnable weight $\alpha$. The task of inner architecture search is learning a set of path weights ${\alpha}$, which can make the supercell achieves best performance. During searching, the path weights are optimized via the gradient descent algorithm. After searching, a compact cell can be obtained by only keeping the top two valuable paths for each node. 

An illustration of the supercell built for each layer is shown in Figure~\ref{fig: inner_search}(a) and corresponding found compact cell is shown in Figure~\ref{fig: inner_search}(b). In Figure~\ref{fig: inner_search}, one supercell in layer $l$ is represented as ${C}_{l}$. ${C}_{l}$ takes the outputs of the previous cell ${C}_{l-1}$ and the cell before previous cell ${C}_{l-2}$ as inputs and outputs ${h}_{l}$. ${C}_{l}$ consists of a sequence of $N$ nodes (N is set to three for clear exposition in Figure~\ref{fig: inner_search}).  Inside ${C}_{l}$, node $i$ takes the two inputs of current cell and the outputs of all previous nodes as inputs. The output of the $i$th node is:
\begin{equation}
\begin{split}
{}& {x}_{l,i}=\sum_{{x}_{j}\in{I}_{l,i}}^{}{{O}_{l}^{j\rightarrow i}({x}_{j})},\\
{}& { O }_{l}^{ j\rightarrow i }({ x }_{ j })=\sum _{ k=1 }^{ S }{ { { \alpha  }_{l}^{ k,  j\rightarrow i } }{ o }^{ k } } ({ x }_{ j }),
\end{split}
\end{equation}
where ${x}_{l,i}$ is the output of the $i$th node in layer $l$. ${I}_{l,i}=\{{h}_{l-1}, {h}_{l-2}, {x}_{l,j<i}\}$ denotes the input set of the $i$th node. ${h}_{l-1}$ and ${h}_{l-2}$ are the outputs of cells ${C}_{l-1}$ and ${C}_{l-2}$.  ${O}_{l}^{j\rightarrow i}$ represents the set of possible layer types. ${\alpha}_{l}^{k
, j\rightarrow i}$ is the weight of operator ${o}^{k}$, which is related to $l$,  as we search for different cell architectures for different layers. While, the cells in the same layer share the same set of continuous variables ${\alpha}$. $\{{o}^{1}, {o}^{2}, \cdots, {o}^{S}\}$ correspond to $S$ possible operations. The output of ${C}_{l}$ is denoted as ${h}_{l}$, which is the concatenation of the outputs of all nodes in ${C}_{l}$ and it can be expressed as:
\begin{equation}
\begin{split}
{h}_{l}={}& {\rm Cell} ({h}_{l-1}, {h}_{l-2})\\
={}& {\rm Concat} \{{x}_{l,i}|i\in\{1, 2, \cdots, N\}\}. 
\end{split}
\end{equation}

\begin{figure}[t]
\setlength{\abovecaptionskip}{0.cm}
\setlength{\belowcaptionskip}{-0.cm}
\centering
\includegraphics[width=3.6in]{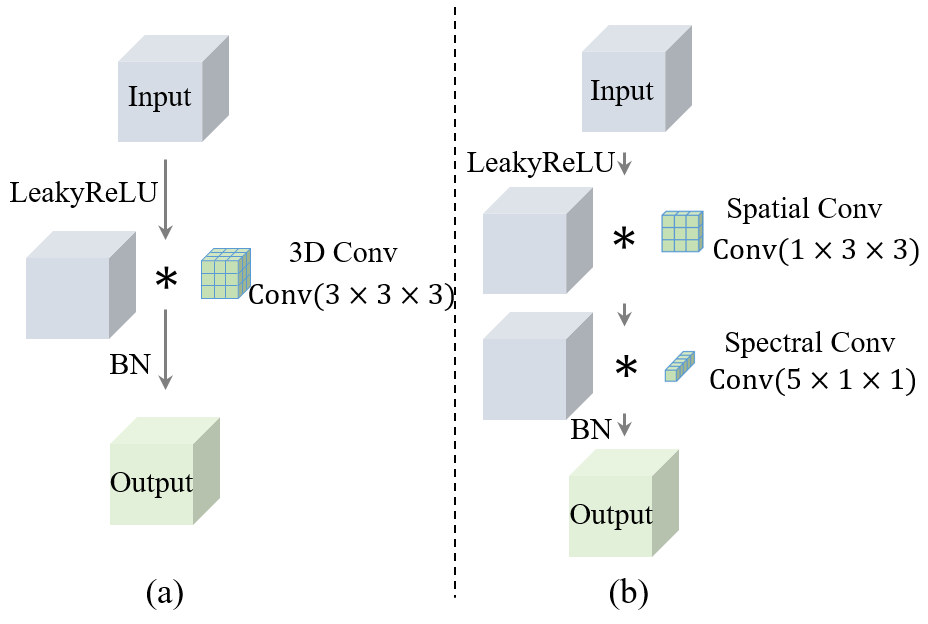}
\caption{3D convolution and 3D asymmetric decomposition convolution. $*$ represents convolution operation. (a)3D convolution. (b) 3D asymmetric decomposition convolution. In 3D asymmetric convolution, the spatial and spectral information is processed separately. The kernel sizes along spatial dimension and spectral dimension can be different (asymmetric)}
\label{fig: asymmetric_conv}
\end{figure}

\vspace{3 pt}
\noindent {\bf 3D Asymmetric Decomposition Convolution}. In this paper, we provide eight different operations, including three common convolution operations, three separable convolution operations, skip connection operation and discarding operation. The eight operations are shown as follows:
\begin{itemize}
    \item con\_3-3: LReLU-$\rm Conv(1\times3\times3)$-$\rm Conv(3\times1\times1)$-BN;
    \item con\_5-3: LReLU-$\rm Conv(1\times5\times5)$-$\rm Conv(3\times1\times1)$-BN;
    \item con\_3-5: LReLU-$\rm Conv(1\times3\times3)$-$\rm Conv(5\times1\times1)$-BN;
    \item Sep\_3-3: LReLU-$\rm Sep(1\times3\times3)$-$\rm Sep(3\times1\times1)$-BN;
    \item Sep\_5-3: LReLU-$\rm Sep(1\times5\times5)$-$\rm Sep(3\times1\times1)$-BN;
    \item Sep\_3-5: LReLU-$\rm Sep(1\times3\times3)$-$\rm Sep(5\times1\times1)$-BN;
    \item skip\_connection: $f(x)=x$;
    \item discarding: $f(x)=0$.
\end{itemize}
where LReLU, BN, Conv and Sep represent LeakyReLU activation function, Batch normalization, common convolution and separable convoluiton. Note that, in this paper, each convolution operation is decomposed into a sequence of two decomposition convolution layers regarding to the spatial dimension and spectral dimension. Figure~\ref{fig: asymmetric_conv} shows an instance, where a 3D convolution with kernel size of $5\times3\times3$ is decomposed into two convolutions with kernel sizes of $1\times3\times3$ and $5\times1\times1$, respectively. Here, we adopt asymmetric decomposition convolution for the following two main reasons: 1) Compared with conventional 3D convolutions, the 3D asymmetric decomposition convolution has a deeper structure and fewer parameters, while having the same receptive field. The advantages of a 3D asymmetric decomposition convolution have been fully demonstrated in~\cite{xie2018rethinking, qiu2017learning, szegedy2016rethinking}; 2) By in-depth analyzing the architectures of HSIs, we find that HSIs have a relatively low spatial resolution and an extremely high spectral resolution. Threrfore, the sizes of kernels along the spatial dimension and spectral dimension should be different. However, increasing the kernel size along the spatial dimension or spectra dimension in common 3D convolution will bring too many parameters, which may affects the final classification performance. By decomposing the common 3D convolution into a pseudo 3D convolution along the spatial dimension and a pseudo 3D convolution along the spectral dimension, it is able to increase the kernel size along the spatial dimension or spectral dimension to get a 3D asymmetric convolution without introducing too much parameters. For instance, a $3\times3\times3$ sized 3D kernel has 27 parameters. Increasing the kernel size along spectral dimension to 5, the number of parameters will increases to $3\time3\times5=45$. The corresponding 3D asymmetric decomposition version has only $3\times3+5=14$ parameters, which is less than one third of the common 3D version. 

\subsection{Outer Widths Search} 

\begin{figure*}[t]
\setlength{\abovecaptionskip}{0.cm}
\setlength{\belowcaptionskip}{-0.cm}
\centering
\includegraphics[width=5.6in]{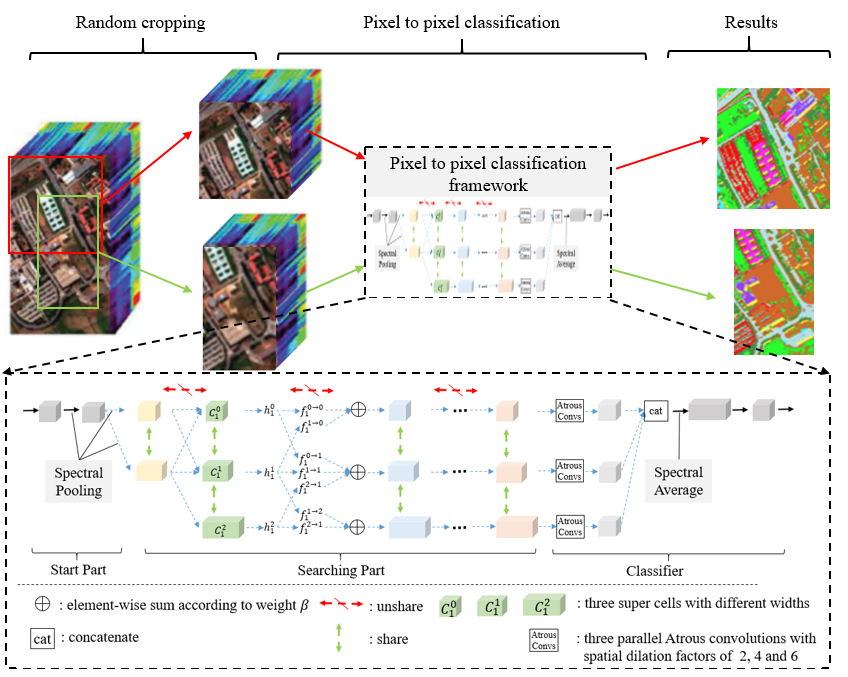}
\caption{The proposed framework. Upper half: pixel-to-pixel classification framework, where patches with arbitrary size is cropped from raw HSI. Then these cropped patches are fed into network to get corresponding classification results. Lower half: the supernet that consists of start part, searching part and classifier part. During searching, the searching part provides three candidate paths with different widths for each layer. After searching, just on path in each layer is retained according to optimized path weight $\beta$.} 
\label{fig: framework}
\end{figure*}

The final architecture is contructed by stacking the found $L$ cells with different widths. Once the topological architectures of the $L$ cells are found via inner architecture search, we need to either heuristically set or automatically search for width for each cell to build the final network. In conventional CNNs, the change of widths is always related to the change of spatial resolutions. For example, doubling the widths of following convolution layers after the spatial resolution of features are reduced. However, to preserve pixel-level information for pixel-to-pixel classification, we do not employ any operations which may reduce the spatial resolution. Thus, the conventional experience of setting the width no longer applies to our case. Aiming to solve this issue, we build an outer search space to automatically search for widths and accomplish the task of setting widths via NAS algorithm.

\vspace{3 pt}
\noindent{\bf Outer Search Space} The outer search space we built for searching for the widths consists of three candidate paths corresponding to different widths, as shown in the bottom half of Figure~\ref{fig: framework}. The complete network has three main parts, including the start part, the searching part and the classifier part. Next, we elaborate on these three parts:
\begin{enumerate}[]

\item Start part, two common 3D convolution layers, each of which followed by a batch normalization layer. In each convolution layer, extracted features are compressed via pooling operation along spectral dimension. In the start part, both 3D convolution layers have 32 channels (width is 32). The kernel size is $(5, 3, 3)$, where the first dimension corresponds to the spectral dimension and the last two dimensions are spatial dimension. The corresponding stride is set to $(2, 1, 1)$.

\item Searching part, including $L$ layers and each layer consisting of three supercells of different widths. This part represents the outer search space. In several classical networks, such as Inception V1 to V4, and MobileNet V3, the topological architecture of different stages are different. In addition, some of the early NAS work search for different architectures for different layers. Inspired by this concept, we search for different cell architectures for different layers. Notably, cells in the same layer share the same architecture.  

\item Classifier, which compresses the outputs of the last layer of searching part to same dimension then concatenate the compressed features to generate concatenated feature. Once the concatenated feature is generated, it is averaged over the spectral dimension. Finally, the averaged feature is fed into a sequence of two 2D convolution layers to get the final classification result.

\end{enumerate}

As shown in the lower part of Figure~\ref{fig: framework}, in the searching part, each layer has three supercells except the first layer which contains two supercells. Three cells at layer $l$ are ${C}_{l}^{0}$, ${C}_{l}^{1}$ and ${C}_{l}^{2}$, widths of which are ${\gamma}^{0}\times W$, ${\gamma}^{1}\times W$ and ${\gamma}^{2}\times W$, where $W$ denote the basic width, which is set to 8 during searching and 16 in the final found network. ${\gamma}^{0}$, ${\gamma}^{1}$ and ${\gamma}^{2}$ are the width variation factors, which are set to 1.0, 1.5 and 2.0 in this paper. The output of each layer can be expressed as:
\begin{equation}
{h}_{l}={{h}_{l}^{0}, {h}_{l}^{1}, {h}_{l}^{2}}
\end{equation}
where ${h}_{l}^{i}$ represents the output of cell ${C}_{l}^{i}$. The width of it is ${\gamma}^{i}WN$, where N denotes the number of nodes in cell ${C}_{l}^{i}$. 

\begin{figure}[h]
\setlength{\abovecaptionskip}{0.cm}
\setlength{\belowcaptionskip}{-0.cm}
\centering
\includegraphics[width=3.5in]{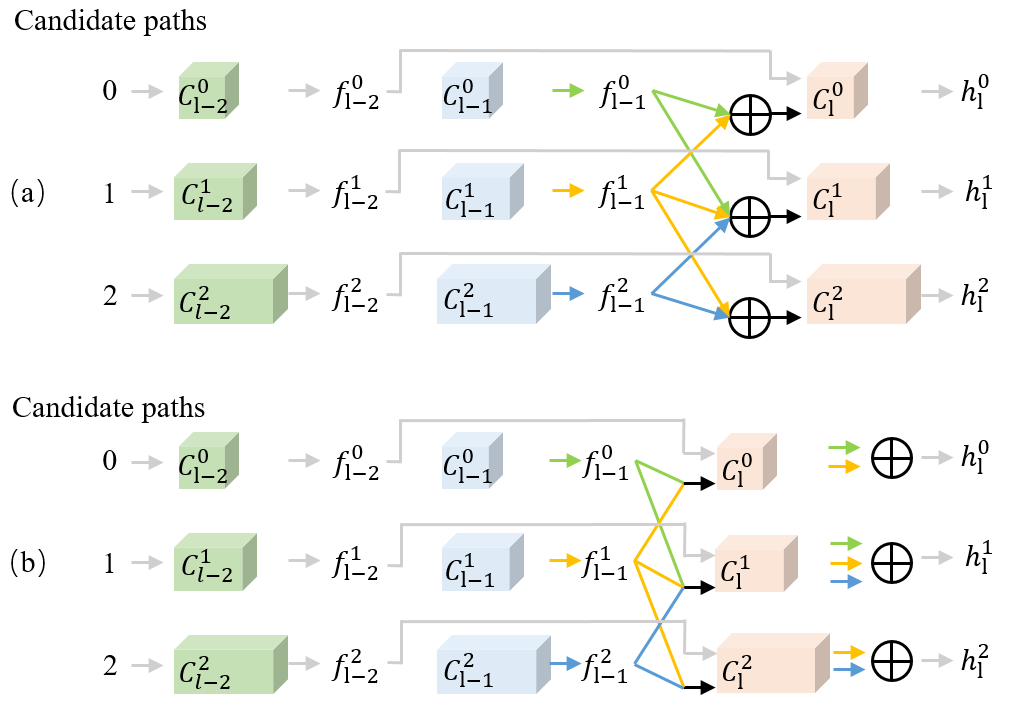}

\caption{Cell sharing strategy. (a) Using cell sharing. Cell ${C}_{l}^{1}$ is shared by features from different candidate paths. During search, ${C}_{l}^{1}$ is called once to process all features from three candidate paths. (b) Without using cell sharing. During search, ${C}_{l}^{1}$ is called for three times to process features from three candidate path, separately.}
\label{fig: cell_sharing}
\end{figure}

\vspace{3 pt}
\noindent{\bf Cell Sharing} As introduced in Section~\ref{Sec: inner_search}, each cell has two inputs regarding to the output of previous cell and the output of cell before the previous cell. In supernet, each layer has three candidate paths. During searching, in layer $l$, the $i$th cell ${C}_{l}^{i}$ also has two inputs: 1) one is ${C}_{l-2}^{i}$, the $i$th cell in layer $l-2$; 2) the other one is chosen from outputs of previous layers, ${C}_{l-1}^{i-1}$, ${C}_{l-1}^{i}$ and ${C}_{l-1}^{i+1}$. As the widths of different cells in the same layers are different, 3D point wise convolution is adopt to map transform outputs ${h}_{l}^{i-1}$, ${h}_{l}^{i}$ and ${h}_{l}^{i-1}$ to ${f}_{l}^{i-1}$, ${f}_{l}^{i}$ and ${f}_{l}^{i+1}$ to match the requirement of cell ${C}_{l}^{i}$. ${f}_{l}^{i-1}$, ${f}_{l}^{i}$ and ${f}_{l}^{i+1}$ have same width, that is ${\gamma}^{i}W$. The output of cell ${C}_{l}^{i}$ is computed with:
\begin{equation}
{h}_{l}^{i}=\sum_{k=i-1}^{i+1}{\beta}_{l}^{k,i}C_l^{k}({f}_{l-1}^{k},{f}_{l-2}^{i}), 
\label{formula: no_share}
\end{equation}
where ${\beta}_{l}^{k,i}$ is the learnable weight of candidate path $k$. During searching, the learnable weights of different candidate paths are updated via gradient descent algorithm, similar with what we do to learn operation weights $\alpha$. From formula~\ref{formula: no_share}, we can see that cell ${C}_{l}^{i}$ is called for three times. In other words, three portions of memory are occupied. Therefore, we propose a cell sharing strategy to save memory and search efficiency. With the proposed strategy, the formula~\ref{formula: no_share} is rewritten as:
\begin{equation}
{h}_{l}^{i}=C_l^i \left(
\sum_{k=i-1}^{i+1}{\beta}_{l}^{k,i}{f}_{l-1}^{k},{f}_{l-2}^{i}
\right),
\label{formula: cell sharing}
\end{equation}
where cell ${C}_{l}^{i}$ is called once. By using cell sharing strategy, the memory consumption is reduced to $1/3$. Figure~\ref{fig: cell_sharing} shows the comparison between using cell sharing and not using cell sharing.. With cell sharing, we first combine the outputs of cell ${C}_{l-1}$ according to corresponding weights then call ${C}_{l}^{i}$ to process the combined output in a single call. Alternatively, cell ${C}_{l}^{i}$ is shared by three candidate paths during the search. The theory behind our proposed cell sharing strategy is similar with meta kernel~\cite{2019Efficient}, where different kernels are summed according to weights then fed into the convolution operation to save memory. Cell sharing can improve our proposed 3D-ANAS in two aspects: 1) cell sharing reduces the demand on computing resources, enabling our 3D-ANAS to be applied to a wider range of domains; 2) cell sharing improves the searching efficiency. As it saves memory consumption, we can use larger batch sizes during the search to increase the search speed. We can also build a deeper and wider supernet to achieve higher accuracy.

Theoretically, outer widths search is an extension of inner architecture search. While, the way to select candidate paths of different widths is different from the way to get the final unit architecture. Here, different layers can not be considered independently. If we just retain the candidate path regarding to the maximum path weight $\beta$ in each layer,  then the widths of adjacent layers in the final network may change drastically. For instance, layer $l-1$ chooses cell ${C}_{l-1}^{0}$ and layer $l$ chooses cell ${C}_{l}^{2}$. According to~\cite{ma2018shufflenet}, this kind of drastic width change has a negative impact on the efficiency. In addition, in our supernet, there is no path between ${C}_{l-1}^{0}$ and ${C}_{l}^{2}$, as illustrated in Figure~\ref{fig: cell_sharing}. Building a new path which does not exist in the search space is unreasonable. To overcome this issue, we regard the β values as probabilities, then use the Viterbi decoding algorithm~\cite{1967Error} to choose candidate path to obtain the final widths search result.

\subsection{pixel-to-pixel Classification Framework}

\begin{figure}[h]
\setlength{\abovecaptionskip}{0.cm}
\setlength{\belowcaptionskip}{-0.cm}
\centering
\includegraphics[width=3.5in]{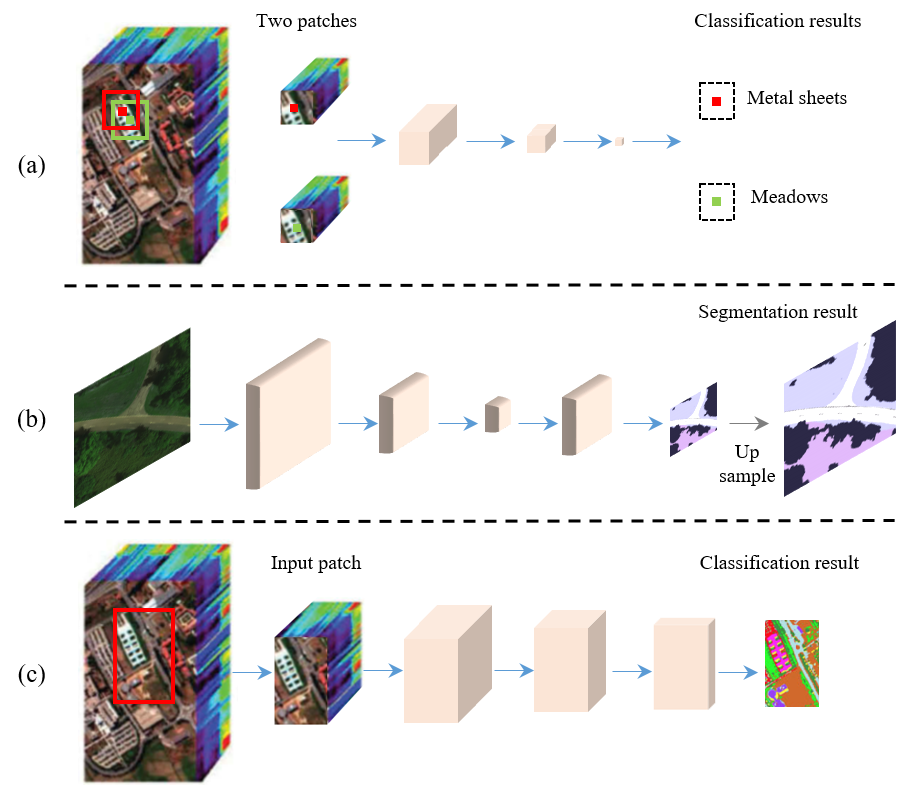}

\caption{Comparison between pixel-to-pixel classification framework, segmentation framework and our pixel-to-pixel classification framework. (a) pixel-to-pixel classification framework; (b) segmentation framework~\cite{long2015fully}; (c) our pixel-to-pixel classification framework.}
\label{fig: framework_comparison}
\end{figure}

Previous spectral-spatial HSI classification approaches usually adopt patch-to-pixel classification framework, where a patch cropped from HSI is fed into trained networks to get the class of a single pixel located in the centre of the cropped patch. In short, all information in a patch is used to classify a single pixel (patch-to-pixel). Two main drawbacks of patch-to-pixel classification framework are: 1) low computing efficiency. The overlap regions of patches cropped for adjacent pixels are repeatedly computed; 2) sensitive to patch size. the  classification  accuracy  is  sensitive  to the patch size. Larger patch  indicates richer contextual  information.  However,  larger  patches  also  represent that the information proportion of the center pixel among pixels in the patch is lower. In addition, different model structures and different HSI datasets have different practical sizes. For instance, the practical patch size is 5 for 3D-CNN model proposed in~\cite{li2017spectral}, while it is 27 for 3D-CNN model proposed in~\cite{chen2016deep}. For SSRN \cite{zhong2018spectral}, the practical patch size is 7. Generally, to obtain the practical patch size, abundant inquiry experiments are carried out.  

In this paper, we adopt a pixel-to-pixel classification framework. Similar with the patch-to-pixel framework, the pixel-to-pixel framework also take patches as input. Differing from the patch-to-pixel framework, pixel-to-pixel framework classifies all pixels in the input patch. In addition, our pixel-to-pixel classification framework is able to process patch with any size. Our pixel-to-pixel framework adopts structures adopted in segmentation tasks. However, HSI classification has higher requirements in pixel-level accuracy compared with segmentation tasks. So our pixel-to-pixel framework discards any spatial resolution reduction operation to preserve pixel-level information. In our pixel-to-pixel framework, the spatial resolution remains unchanged from the input layer to the final prediction result. Meanwhile, to reduce the amount of computations, spectral pooling is employed to condense features along the spectral dimension. Figure~\ref{fig: framework_comparison} illustrates the difference between the patch-to-pixel framework, the segmentation framework and our pixel-to-pixel framework. 

\subsection{Training strategy}
\label{sec: training strategy}

\begin{figure*}[h]
\setlength{\abovecaptionskip}{0.cm}
\setlength{\belowcaptionskip}{-0.cm}
\centering
\includegraphics[width=6.0in]{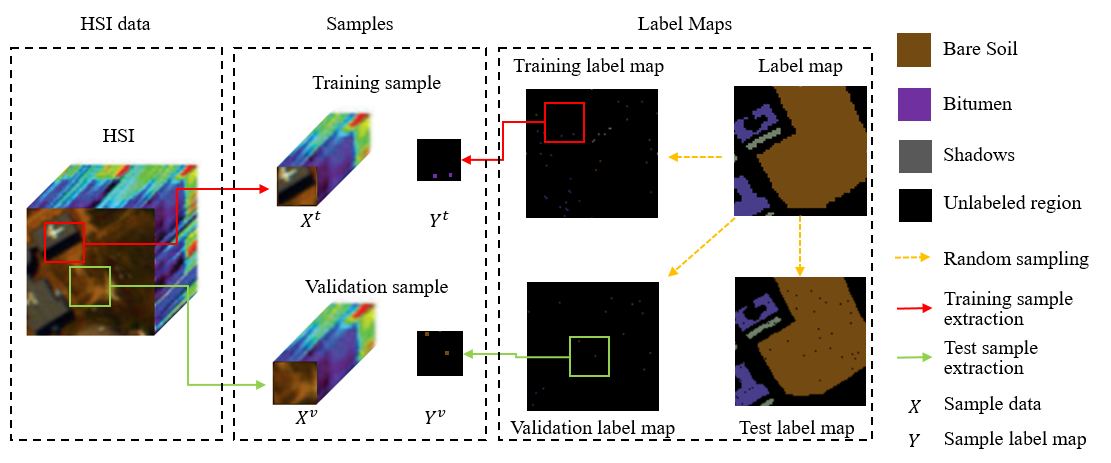}

\caption{An illustration of sample extraction. Label map is randomly divided into three sub-maps, including training label map, validation label map and test label map. To generate a training sample, we first randomly crop a ${W}_{1}\times {W}_{2}$-sized patch with sparse labeling information as the label of the training sample (${Y}^{t}$). Then, according to the spatial location of ${Y}^{t}$, we correspondingly crop a ${W}_{1}\times {W}_{2} \times L$-sized 3D cuboid from HSI as the data of training sample (${X}^{t}$). We extract validation samples in the same way except that label patch are extracted from validation label map.}
\label{fig: sample_extraction}
\end{figure*}

As the basic framework we used in this paper is a pixel-to-pixel classification framework, which is different from the patch-to-pixel classification framework used in previous works. The training samples used in this paper are also different from previous work. Next, we explain how to extract samples and train our pixel-to-pixel framework. An labelled sample contains two parts, input sample data $X$ and corresponding label $Y$. In the patch-to-pixel framework, $X$ is a 3D cuboid. For instance, a $W\times W\times L$-sized cuboid, where $W$ is the patch size and $L$ denotes the number of spectral bands. The corresponding label $Y$ is a single integer. In our pixel-to-pixel classification framework, input data $X$ is still a 3D cuboid. For example, a ${W}_{1}\times {W}_{2}\times L$-sized cube, where ${W}_{1}\times {W}_{2}$ are spatial resolution and ${W}_{1}$ and ${W}_{2}$ can be different. The corresponding label $Y$ is a ${W}_{1}\times {W}_{2}$-sized 2D label map. In HSI dataset, not every pixel is labelled, so the label map $Y$ is sparse. The procedure of sample extraction is shown in Figure~\ref{fig: sample_extraction}. 

Our proposed 3D-ANAS has two optimizing stages, including 1) architecture searching stage (supernet optimizing stage) and 2) the optimizing stage of the final network. 

In the searching stage, we randomly generated 400 training samples. Then we randomly distributed training samples into two subsets ${Train}_{\theta}$ and ${Train}_{A}$ , each of which has 200 training samples. ${Train}_{\theta}$ is used to optimize superent parameters $\theta$, which are learnable parameters in convolution layers and batch normalization layers. ${Train}_{A}$ is used to update the architecture parameters $\alpha$ and $\beta$. The optimization function of searching process can be expressed as:
\begin{equation}
\mathop{\arg\min}_{\alpha, \beta, \theta} {\rm CELoss}({F}_{s}(x,y|\alpha, \beta, \theta)),
\end{equation}
where CEloss is cross entropy loss, ${F}_{s}$ denotes supernet, $x$ and $y$ are the sample data and the sample label. During searching stage, architecture parameters ($\alpha$ and $\beta$) and supernet parameters ($\theta$) are alternately updated. To evaluate the performance of supernet, we build a validation set $Val$. Specifically, based on sliding windows algorithm, we crop a sequence of patches from validation label map to build validation set. We evaluate the supernet for every epoch and retain the best one. At the end of searching stage, we deduce the final network architecture according to the retained supernet. 

In the final network optimizing stage, we randomly generate training samples in each iteration and evaluate the found network for every 100 iterations. We stop the optimization process when the classification accuracy on validation set no longer increase and the loss on validation set no longer decreases. The optimization function of this stage is:
\begin{equation}
\mathop{\arg\min}_{\theta} {\rm CELoss}({F}_{f}(x,y|\theta)),
\label{equ: final optimization function}
\end{equation}
where ${F}_{f}$ is the final network. 

\section{Experimental Results}

\subsection{Data Description and Experiment Design}
In this section, we organize three groups of experiments. Firstly, to evaluate the performance of our proposed 3D-ANAS, we carry out experiments on three different HSI datasets which are collected after 2001 and compare the proposed 3D-ANAS with other 3D-CNN based and NAS based methods. Secondly, to better comprehend the architectures found by 3D-ANAS, we conduct architecture analysis experiments. Finally, we present an ablation study to analyze the effectiveness of each component in 3D-ANAS and the inference speed of architectures found by 3D-ANAS.

\begin{figure*}[h]
\setlength{\abovecaptionskip}{0.cm}
\setlength{\belowcaptionskip}{-0.cm}
\centering
\includegraphics[width=7.0in]{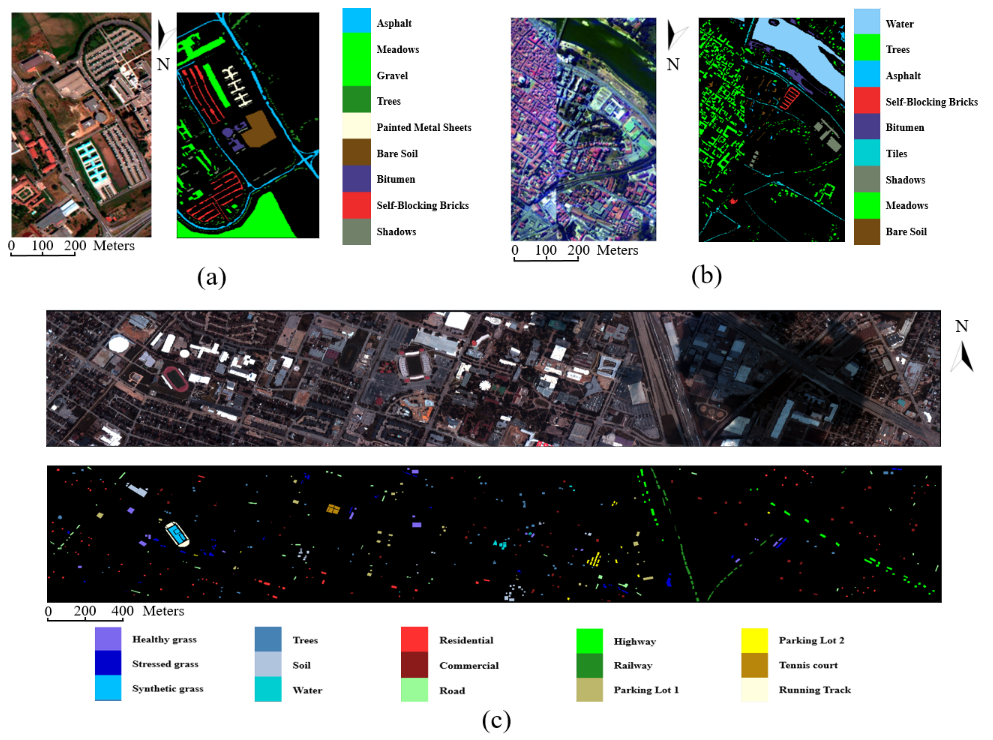}
\caption{False color composites and ground truth maps of datasets. (a) Pavia University; (b) Pavia Centre; (c) Houston University}
\label{fig: datasets}
\end{figure*}

\begin{table*}[!t]
\setlength{\abovecaptionskip}{0.cm}
\setlength{\belowcaptionskip}{-0.cm}
\renewcommand{\arraystretch}{1.3}
\caption{Sample distribution information of datasets}
\label{tab: datasets}
\centering
\begin{tabular}{c|c|c|c|c|c|c}
\hline
  & \multicolumn{2}{c|}{Pavia University} & \multicolumn{2}{c|}{Pavia Centre} & \multicolumn{2}{c}{Houston University} \\
\hline
No. & Category & Labelled samples & Category & Labelled samples & Category & Labelled samples \\
1	& Asphalt              &  6631  & Water                & 824  & Healthy Grass   &  1251 \\  
2	& Meadows              &  18649 & Trees                & 820  & Stressed Grass  &  1254 \\
3	& Gravel               &  2099  & Asphalt              & 816  & Synthetic Grass &  697  \\
4	& Trees                &  3064  & Self-Blocking Bricks & 808  & Trees           &  1244 \\
5	& Painted Metal Sheets &  1345  & Bitumen              & 808  & Soil            &  1242 \\
6	& Bare Soil            &  5029  & Tiles                & 1260 & Water           &  325  \\
7	& Bitumen	           &  1330  & Shadows              & 476  & Residential     &  1268 \\
8   & Self-Blocking Bricks &  3682  & Meadows              & 824  & Commercial      &  1244 \\
9	& Shadows              &  947   & Bare Soil            & 820  & Road            &  1252 \\
10  &                      &        &                      &      & Highway         &  1227 \\
11  &                      &        &                      &      & Railway         &  1235 \\
12  &                      &        &                      &      & Parking Lot 1   &  1233 \\
13  &                      &        &                      &      & Parking Lot 2   &  469  \\
14  &                      &        &                      &      & Tennis Court    &  428  \\
15  &                      &        &                      &      & Running Track   &  660  \\
\hline
    & Total                &  42776 & Total                & 7456 & Total           & 15029 \\
\hline
\end{tabular}
\end{table*}

\subsubsection{Data Description} 
The three HSI datasets used in this paper are Pavia University, Pavia Centre and Houston University. The first two datasets were collected by Reflective Optics System Imaging Spectrometer (ROSIS) sensor in 2001, during a flight campaign over Pavia, nothern Italy. After correction, Paiva University has 103 spectral bands and Paiva Centre retains 102 spectral bands. Pavia University covers 610$\times$340 pixels and Pavia Centre consists of 1096$\times$715 pixels. Both Pavia University and Pavia Centre have 9 classes of ground objects. The geometric resolution of these two datasets is 1.3 meters. The last dataset Houston University was captured by ITRES-CASI 1500 hyperspectral Imager on June, 2012. It covers the regions over the University of Houston campus and the neighboring urban area. Houston University has 144 spectral bands, covering the range from 0.36 to 1.05 $\upmu$m. It consists of 349$\times$1905 pixels with a high spatial resolution of 2.5 m. 15 land-cover classes are included in this dataset. Sample distribution information is listed in Table~\ref{tab: datasets} and shown in Figure~\ref{fig: sample_extraction}.

\begin{figure*}[h]
\setlength{\abovecaptionskip}{0.cm}
\setlength{\belowcaptionskip}{-0.cm}
\centering
\includegraphics[width=6.0in]{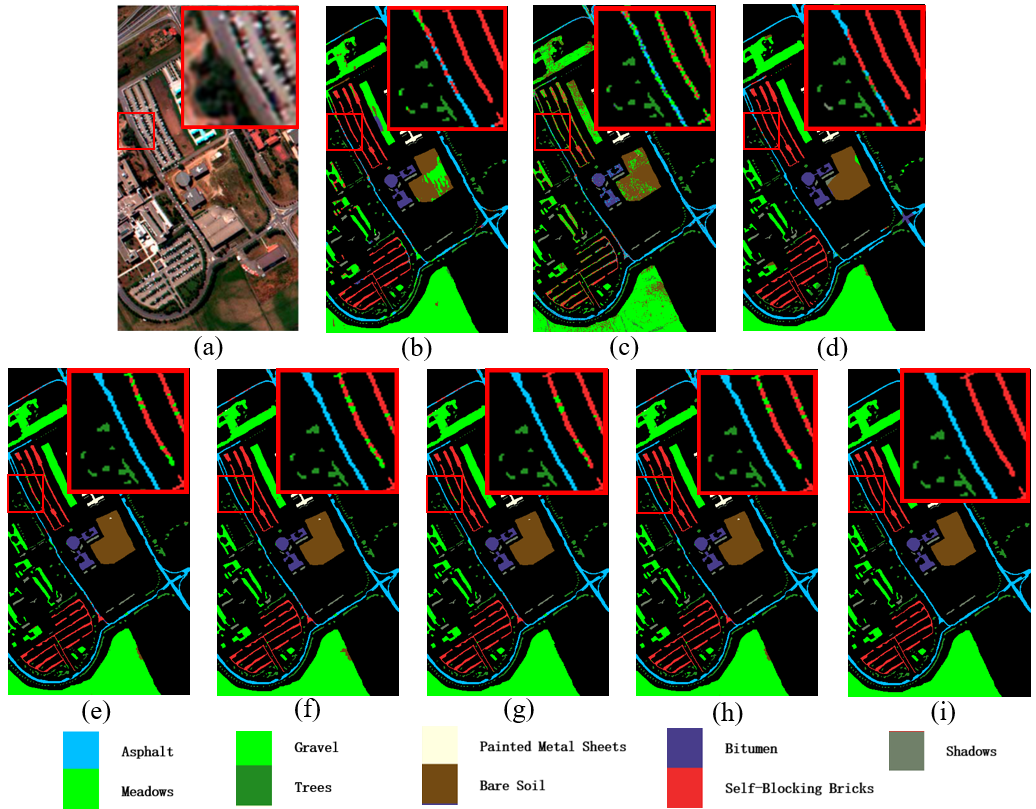}
\caption{Comparison experimental results on Pavia University using 30 training samples each category. (a) False color composite; (b) 3D-LWNet OA=89.25\%; (c) 1-D Auto-CNN, OA=81.75\%; (d) 3-D Auto-CNN, OA=93.36\%; (e)3D-ANAS, OA=97.81\%; (f) 3D-ANAS+MS, OA=97.69\%; (g)3D-ANAS+OV, OA=97.94\%; (h)3D-ANAS+MS+OV, OA=97.92\%;}
\label{fig: PaviaU_30}
\end{figure*}

\subsubsection{Implementation Details}
As we introduced in Section~\ref{sec: training strategy}, the proposed 3D-ANAS has two optimizing phases, including the architecture searching phase and and the final network optimizing phase. In this section, we first expound implementation details of these two stages in detail, then we introduce experiment design.

\vspace{3 pt}
\noindent{\bf Architecture searching}. For three datasets employed in this paper, we build three supernets, each of which has two layers of supercells. Each supercell contains three nodes. The task of architecture searching is optimizing supernets to get optimal weights of different operations and candidate paths. We crop patches with spatial resolution of 32$\times$32 as training and validation samples. We train each supernet on a single GTX1080Ti GPU, which has 11G memory. For Pavia University, Pavia Centre and Houston University, we set batch size to 5, 5 and 4, respectively. We use Adam optimizer to optimize architecture parameters ($\alpha$ and $\beta$), where both learning rate and weight decay are 0.001. We employ standard SGD optimizer to update supernet parameters ($\theta$), where momentum and weight decay are set to 0.9 and 0.0003, respectively. The learning rate decays from 0.025 to 0.001 according to cosine annealing strategy~\cite{loshchilov2016sgdr}. We training the supernet at most 100 epochs. The first 20 epochs is warm up stage, where we only optimize $\theta$. From epoch 21, we update architecture parameters and supernet parameters alternately in each iteration. 

\vspace{3 pt}
\noindent{\bf Final network optimizing}. For all three final networks found for three datasets, same optimizing setting are used. Similar with the setup in the searching stage, we still crop 32$\times$32$\times$L-sized patches as training and validation samples. In the final network, we only need to update network parameters, as shown in formula~\ref{equ: final optimization function}. The optimizer used here is SGD optimizer. The batch size is set to 12. The initial learning rate is set to 0.1. It is adjusted based on poly learning rate policy with power of 0.9 ($lr=init\_lr\cdot({1-\frac{iter}{max\_iter}}^{power})$). We optimize the final network at most 600k iterations and evaluate the network on validation set for every 100 iterations. This training process stops once the performance and loss of the validation set reach saturation. We use random crop, horizontal and vertical flipping as well as random rotations$\in \{{0}^{\circ}, {90}^{\circ}, {180}^{\circ}, {270}^{\circ}\}$ as data augmentation.

\vspace{3 pt}
\noindent{\bf Inference Setting}. For our pixel-to-pixel classification framework, we employ two inference augmentation strategies to further improve the performance. 1) Multi-scale inference (termed as MS). As our 3D-ANS can accept patches of arbitrary spatial size as input, we crop the patches in three different scales, $16\times16\times L$, $32\times32\times L$ and $48\times48\times L$, respectively. The final prediction result is obtained by averaging prediction results of three scales. 2) Overlap inference (termed as OV). During inference, we crop patches in sliding window strategy then feed the cropped patches to trained networks. In the sliding window strategy, there are two hyper parameters, which are stride and window size. When stride is set to half of the window size, the adjacent patches have a overlap region, which is processed twice. In overlap inference, we set stride to half of the window size and take the average result as the final prediction result. Based on the two inference augmentation strategies, we implement the inference process with four different strategies: 1) 3D-ANAS, which uses a single scale 32 32 for inference and patches without overlapping regions; 2) 3D-ANAS+MS,  which uses of the MS strategy for inference; 3) 3D-ANAS+OV, inference using OV strategy; 4) 3D-ANAS+MS+OV, inference using both MS and OV strategies.

\vspace{3 pt}
\noindent{\bf Experiment Design}. To demonstrate the effectiveness of the proposed 3D-ANAS, we distribute the labelled samples in two different proportions: 1) Distribution 1, we randomly select 20 labelled pixels from each category to build training label map and randomly select 10 labelled pixels from each category to build validation label map. Then the rest are reserved as test label map. Taking Pavia University as an example, training label map contains 180 labelled pixels, validation label map consists of 90 labelled pixels and test label map has 42476 labelled pixels. 2) Distribution 2, we randomly choose 30 labelled pixels and 10 labelled pixels from each category to build training label map and validation label map, respectively. As samples are randomly distributed, we run each experiment for ten times and compare the average result to ensure the comparisons are fair and stable.

\subsection{Comparison Experiments}

In this section, we focus on comparing our proposed 3D-ANAS with other CNN based HSI classification methods. Here, three comparison methods published recently are employed. We conduct 3D-LWNet based on the published code\footnote{\url{https://github.com/hkzhang91/LWNet}}. The results of 1-D Auto-CNN and 3-D Auto-CNN are obtained by using the published source code\footnote{\url{https://github.com/YushiChen/Auto-CNN-HSI-Classification}}. Tables~\ref{tab: PaviaU_20}-\ref{tab: Houston_30} list the comparison results and Figures~\ref{fig: PaviaU_30}-\ref{fig: Houston_30} show the corresponding visualiztion results. 

The results of comparison experiments on Paiva University are listed in Table~\ref{tab: PaviaU_20} and Table ~\ref{tab: PaviaU_30}. The corresponding visual comparison results are shown in Figure~\ref{fig: PaviaU_30}. From the comparison results we can draw three conclusions: 1) compared with 1D CNN based methods, 3D CNN based methods generally have better performance. Among these methods compared in Tables\ref{tab: PaviaU_20}-\ref{tab: PaviaU_30}, the classification accuracy of 1-D Auto-CNN is lower that others. The is consistent with the statements in recent spectral-spatial HSI classification works. Using both spectral and spatial information is benefit for improving classification accuracy. 2) our proposed 3D-ANAS achieves much better performance compared with other CNN model based methods and NAS based approaches. For instance, with using 20 training samples per category, 3D-ANAS achieves 96.53\% OA, 96.31\% AA and 95.39\% K, which are 5.37, 5.59, and 6.88 percentage points higher than 3D Auto-CNN, respectively. 3) Our inference augmentation strategy can further improve the performance.

\newcolumntype{C}[1]{>{\centering\let\newline\\\arraybackslash\hspace{0pt}}m{#1}}
\begin{table}[]
\setlength{\abovecaptionskip}{0.cm}
\setlength{\belowcaptionskip}{-0.cm}
\renewcommand{\arraystretch}{1.3}
\caption{Comparison Experimental Results on Pavia University Using 20 Training Samples Each Category}
\label{tab: PaviaU_20}
\centering
\begin{tabular}{C{0.8cm}|C{0.7cm}C{0.65cm}C{0.65cm}|C{0.65cm}C{0.65cm}C{0.65cm}C{0.65cm}}
\hline
Models & 3D-LWNet & 1-D Auto-CNN & 3-D Auto-CNN & 3D-ANAS & 3D-ANAS +MS & 3D-ANAS +OV & 3D-ANAS +MS+OV \\

1      &  82.43   &  69.69         &  88.24      &    93.21    &  \textit{93.35}      &    \textbf{93.39}   &   93.15  \\  
2      &  84.76   &  76.37         &  90.72      &    99.1     &  99.04      &    \textit{99.21}   &   \textbf{99.43}  \\ 
3      &  76.88   &  73.43         &  92.11      &    99.28    &  99.32      &    \textit{99.52}   &   \textbf{99.71}  \\ 
4      &  91.45   &  90.2          &  81.27      &    92.22    &  \textbf{93.05}      &    \textit{92.65}   &   \textit{92.65}  \\ 
5      &  96.23   &  \textit{96.54}         &  93.12      &    \textbf{100}      &    \textbf{100}      &    \textbf{100}     &   \textbf{100} \\ 
6      &  92.50   &  75.48         &  \textbf{98.47}      &    97.36    &  97.56      &   98.16    &   \textit{98.18}  \\ 
7      &  93.56   &  88.83         &  96.14      &    99.62    &  \textit{99.69}      &    \textbf{100}     &   \textbf{100} \\ 
8      &  \textit{96.01}   &  77.59         &  \textbf{96.84}      &    87.35    &  88.12      &    87.68   &   87.92  \\ 
9      &  89.16   &  96.66         &  79.62      &    98.69    &  \textbf{99.35}      &    \textit{99.13}   &   \textit{99.13}  \\ 
\hline
OA     &  87.10   &  77.65         &  91.16      &    96.53    &  96.69      &    \textit{96.79}   &   \textbf{96.88}  \\ 
AA     &  89.22   &  82.75         &  90.72      &    96.31    &  96.61      &    \textit{96.64}   &   \textbf{96.69}  \\ 
K      &  83.35   &  71.51         &  88.51      &    95.39    &  95.61      &    \textit{95.73}   &   \textbf{95.85}  \\ 
\hline
\end{tabular}
\end{table}

\begin{table}[!t]
\setlength{\abovecaptionskip}{0.cm}
\setlength{\belowcaptionskip}{-0.cm}
\renewcommand{\arraystretch}{1.3}
\caption{Comparison Experimental Results on Pavia University Using 30 Training Samples Each Category}
\label{tab: PaviaU_30}
\centering
\begin{tabular}{C{0.8cm}|C{0.7cm}C{0.65cm}C{0.65cm}|C{0.65cm}C{0.65cm}C{0.65cm}C{0.65cm}}
\hline
Models & 3D-LWNet & 1-D Auto-CNN & 3-D Auto-CNN & 3D-ANAS & 3D-ANAS +MS & 3D-ANAS +OV & 3D-ANAS +MS+OV \\
1 &  82.32    &  76.43         & 89.70       & 93.21      & 93.3      & \textit{93.32}  & \textbf{93.38}     \\  
2 &  88.84    &  81.9          & 97.92       & \textbf{98.5}       & 98.22     & \textit{98.46}  & 98.32   \\ 
3 &  83.74    &  74.04         & 92.31       & \textit{99.61}      & \textit{99.61}     & \textbf{99.81} &  \textbf{99.81} \\ 
4  &  \textbf{94.11}   &  \textit{93.38}         & 71.22       & 98.09      & 98.02     & \textit{98.38} & 98.35 \\ 
5 &  \textit{96.90}    &  96.13         & 95.12       & \textbf{100}        & \textbf{100}       & \textbf{100}   & \textbf{100}    \\ 
6  &  92.10   &  81.81         & 96.96       & 99.8       & \textit{99.82}     & \textbf{99.92} & \textbf{99.92}   \\ 
7  &  95.46   &  88.23         & 95.99       & \textbf{99.77}      & \textit{99.54}     & \textit{99.54}   & \textbf{99.77}    \\ 
8 &  93.19    &  74.05         & 94.98       & 96.55      & 96.5      & \textit{97.65}   & \textbf{98.03}    \\ 
9 &  92.43    &  \textit{95.65}         & 80.64       & \textbf{100}        &  \textbf{100}      & \textbf{100}     &  \textbf{100} \\ 
\hline
OA   &  89.25 &  81.75         & 93.36       & 97.81      & 97.69       &  \textbf{97.94} &  \textit{97.92}    \\ 
AA   &  91.01 &  84.62         & 90.54       & 98.39      &  98.34      &  \textit{98.56} &  \textbf{98.62} \\ 
K  &  86.05   &  76.55         & 91.50       & 97.1       &  96.94      &  \textbf{97.27} &   \textit{97.25}    \\ 
\hline
\end{tabular}
\end{table}

\begin{figure*}[]
\setlength{\abovecaptionskip}{0.cm}
\setlength{\belowcaptionskip}{-0.cm}
\centering
\includegraphics[width=6.0in]{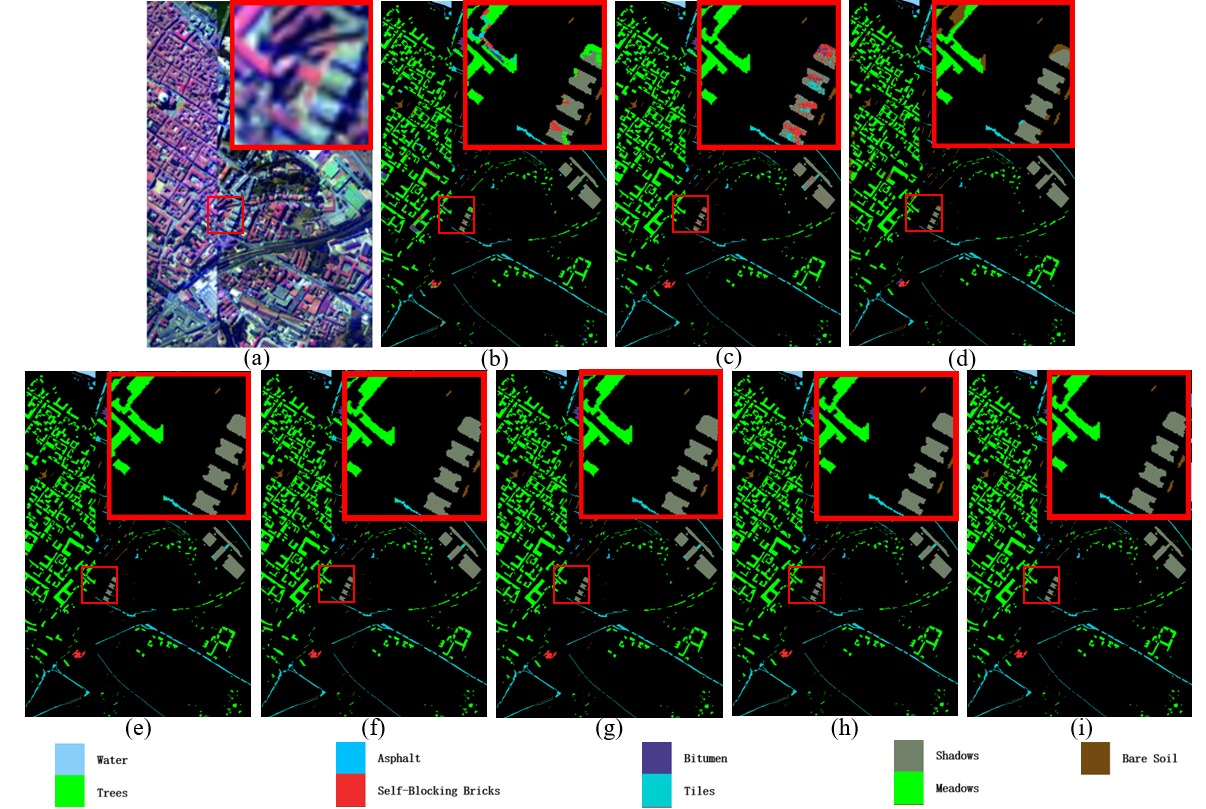}
\caption{Comparison experimental results on Pavia Centre using 30 training samples each category. (a) False color composite; (b) 3D-LWNet OA=94.22\%; (c) 1-D Auto-CNN, OA=96.79\%; (d) 3-D Auto-CNN, OA=96.99\%; (e)3D-ANAS, OA=99.47\%; (f) 3D-ANAS+MS, OA=99.49\%; (g)3D-ANAS+OV, OA=99.51\%; (h)3D-ANAS+MS+OV, OA=99.50\%;}
\label{fig: Pavia_30}
\end{figure*}

\begin{figure*}[!h]
\setlength{\abovecaptionskip}{0.cm}
\setlength{\belowcaptionskip}{-0.cm}
\centering
\includegraphics[width=6.6in]{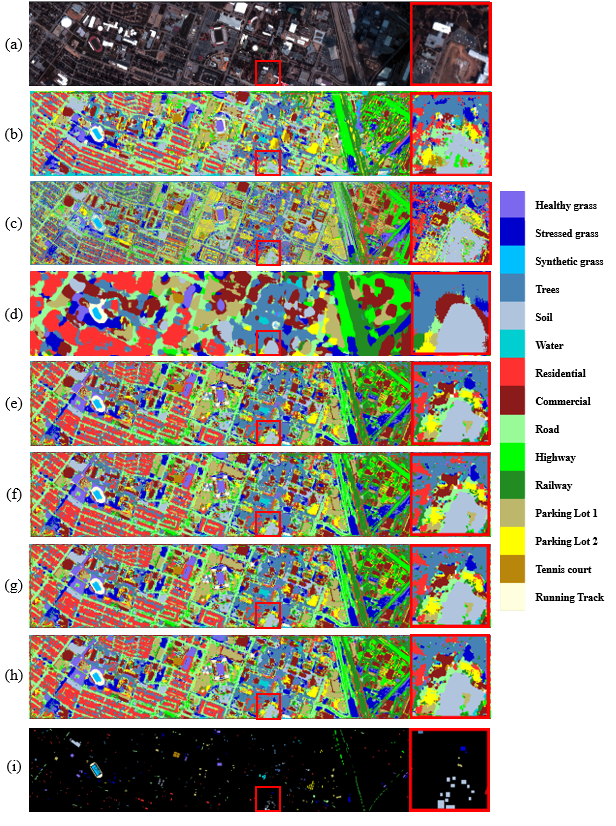}
\caption{Comparison experimental results on Houston University using 30 training sample each category. (a) False color composite; (b) 3D-LWNet OA=84.17\%; (c) 1-D Auto-CNN, OA=60.36\%; (d) 3-D Auto-CNN, OA=84.45\%; (e)3D-ANAS, OA=86.16\%; (f) 3D-ANAS+MS, OA=86.51\%; (g)3D-ANAS+OV, OA=87.67\%; (h)3D-ANAS+MS+OV, OA=87.82\%;}
\label{fig: Houston_30}
\end{figure*}

\begin{table}[]
\setlength{\abovecaptionskip}{0.cm}
\setlength{\belowcaptionskip}{-0.cm}
\renewcommand{\arraystretch}{1.3}
\caption{Comparison Experimental Results on Pavia Centre Using 20 Training Samples Each Category}
\label{tab: Pavia_20}
\centering
\begin{tabular}{C{0.8cm}|C{0.7cm}C{0.65cm}C{0.65cm}|C{0.65cm}C{0.65cm}C{0.65cm}C{0.65cm}}
\hline
Models & 3D-LWNet & 1-D Auto-CNN & 3-D Auto-CNN & 3D-ANAS & 3D-ANAS +MS & 3D-ANAS +OV & 3D-ANAS +MS+OV \\
1    &  99.61    &   \textit{99.81}       &   99.56        & \textbf{100}       &  \textbf{100}      &   \textbf{100}     &  \textbf{100}       \\  
2    &  91.85    &   80.9       &    87.79       &  \textbf{94.66}      &   \textit{94.64}     &   \textbf{94.66}     & 94.61        \\ 
3    &  86.77    &   89.04       &   82.98        &  \textit{96.21}      &    95.59    &  \textbf{96.34}      & 96.11         \\ 
4    &  92.85    &   73.53       &   98.85        &  99.89      &    \textit{99.96}    &  \textbf{100}      &  \textbf{100}       \\ 
5    &  95.53    &   90.05       &   95.83        &  97.6      &     97.77   &    \textit{98.79}    &    \textbf{98.96}     \\ 
6    &   80.97   &   95.53       &   93.22        &  99.8      &     99.86   &   \textit{99.91}     &     \textbf{99.97}    \\ 
7	 &   85.66   &   84.23       &   94.94        &  97.88      &    97.89    &   \textit{97.92}      &   \textbf{97.99}      \\ 
8    &   85.06   &   98.17       &   95.12        &  \textit{99.25}      &    \textit{99.25}    &   \textbf{99.26}      &   99.23      \\ 
9    &   91.34   &   98.89       &   85.29        &  \textit{99.96}      &    \textbf{100}    &    \textbf{100}    &    \textbf{100}     \\ 
\hline
OA   &   92.42   &  96.18        &   96.25         & 99.21        &    99.2     &   \textbf{99.28}     &   \textit{99.27}        \\ 
AA   &   89.96   &  90.02        &   92.62         & \textit{98.36}        &     98.33   &   \textbf{98.54}     &   \textbf{98.54}    \\ 
K    &   89.41   &  94.6         &   94.71        &  98.88        &    98.87    &   \textbf{98.98}     &   \textit{98.97}  \\ 
\hline
\end{tabular}
\end{table}

\begin{table}[]
\setlength{\abovecaptionskip}{0.cm}
\setlength{\belowcaptionskip}{-0.cm}
\renewcommand{\arraystretch}{1.3}
\caption{Comparison Experimental Results on Pavia Centre Using 30 Training Samples Each Category}
\label{tab: Pavia_30}
\centering
\begin{tabular}{C{0.8cm}|C{0.7cm}C{0.65cm}C{0.65cm}|C{0.65cm}C{0.65cm}C{0.65cm}C{0.65cm}}
\hline
Models & 3D-LWNet & 1-D Auto-CNN & 3-D Auto-CNN & 3D-ANAS & 3D-ANAS +MS & 3D-ANAS +OV & 3D-ANAS +MS+OV \\
1   &   99.52     &   99.76        &   99.78     &   99.94     &   99.95     &  \textit{99.96}  &  \textbf{99.97}   \\  
2   &   93.92     &    87.9       &   92.48     &   95.49     &   \textit{95.56}     &   \textbf{95.59}   &  95.51 \\ 
3   &   89.26     &    91.16       &   85.81     &   96.73     &   \textbf{97.09}     &   \textit{96.86}  &  96.57  \\ 
4   &  91.1       &    79.72       &   97.32     &    99.62    &   \textit{99.77}    &   \textbf{99.81}  &   \textbf{99.81} \\ 
5   &   96.09     &    92.1       &   97.02     &   \textbf{99.69}     &   \textit{99.59}     &   99.5   &  99.51 \\ 
6   &   90.73     &    96.47       &    95.91    &   99.35     &   99.40     &   \textit{99.65}   & \textbf{99.73}  \\ 
7	&   93.24     &    85.43       &   95.29     &   97.55     &   \textit{97.57}     &   \textbf{97.60}   & \textbf{97.60}  \\ 
8   &    87.32    &   97.88        &   95.13     &   99.9     &   \textit{99.92}     &   \textbf{99.93}   &  \textbf{99.93} \\ 
9   &    93.7     &    \textit{98.58}       &   91.84     &   \textbf{100.0}     &   \textbf{100.0}     &  \textbf{100.0}    &  \textbf{100.0} \\ 
\hline
OA  &    94.22    &    96.79       &   96.99     &   99.47     &   99.49     &   \textbf{99.51}  &  \textit{99.50}  \\ 
AA  &    92.76    &    92.11       &   94.51     &   98.70     &    \textit{98.76}    &   \textbf{98.77}   &  98.74 \\ 
K   &    91.92    &    95.47       &    95.76    &   99.24     &    \textit{99.28}    &   \textbf{99.30}   &  \textbf{99.30} \\ 
\hline
\end{tabular}
\end{table}

As shown in Table~\ref{tab: PaviaU_20}, using MS improves OA, AA and K by 0.16, 0.30 and 0.22 percentage points, respectively. Using OV improves OA, AA and K by 0.26, 0.33 and 0.34 percentage points, respectively. Compared with 3D-ANAS which does not employ any inference augmentation strategies,  3D-ANAS+MS+OV which uses both MS and OV strategies improves OA, AA and K by 0.35, 0.38 and 0.46 percentage points, respectively. In addition, it also demonstrates that that OV inference augmentation strategy is more useful than MS inference augmentation strategy in improving classification accuracy. From Table~\ref{tab: PaviaU_20} to Table~\ref{tab: PaviaU_30}, with number of training samples per class increases from 20 to 30, all methods obtain much higher classification accuracy. For instance, the OA of 3D-LWNet increases from 87.10\% to 89.25\%. (With transferring model from Salinas, 3D-LWNet achieves 93.46\% OA, 93.63\% AA and 91.43\% K, which are highly competitive with 3-D Auto-CNN. Nevertheless, as other methods does not use any transfer learning strategies, here we show the results of 3D-LWNet without using cross-sensor and cross-modal transfer learning strategies for fair comparison.) Our proposed 3D-ANAS still has better performance compared with others. Unexpectedly, in Table~\ref{tab: PaviaU_30}, using MS during inference slightly degrades the final performance. We conjecture this because we only use a single scale during training. To save space, instead of showing both corresponding results of Table~\ref{tab: PaviaU_20} and Table~\ref{tab: PaviaU_30}, we illustrate the visual results of using 30 training samples each category in Figure~\ref{fig: PaviaU_30}. For clearly showing the difference, we place a local enlarged patch in the top right corner of each result map. From the local enlarged patches, we can see that that fewer pixels were misclassified in the results of 3D-ANAS, 3D-ANAS+MS, 3D-ANAS+OV, and 3D-ANAS+MS+OV than in the other methods. In the local enlarged region, part of Asphalt pixels (class 1, cyan) is misclassified as Self-Blocking Bricks (class 8, red) by 3D-LWNet and 3-D Auto-CNN, most of the pixels that belong to Self-Blocking Bricks are misclassified as Meadows (class 2, green) by 1-D Auto-CNN. In contrast to this, in the results of 3D-ANAS, all pixels of Asphalt are correctly classified and a small number of pixels belonging to Self-Blocking Bricks are misclassified.

\begin{table}[!t]
\setlength{\abovecaptionskip}{0.cm}
\setlength{\belowcaptionskip}{-0.cm}
\renewcommand{\arraystretch}{1.3}
\caption{Comparison Experimental Results on Houston University Using 20 Training Samples Each Category}
\label{tab: Houston_20}
\centering
\begin{tabular}{C{0.8cm}|C{0.7cm}C{0.65cm}C{0.65cm}|C{0.65cm}C{0.65cm}C{0.65cm}C{0.65cm}}
\hline
Models & 3D-LWNet & 1-D Auto-CNN & 3-D Auto-CNN & 3D-ANAS & 3D-ANAS +MS & 3D-ANAS +OV & 3D-ANAS +MS+OV \\
1   &   79.16     &    69.98       &  \textbf{84.01}      &   82.06     &  82.47      &   82.64  &  \textit{83.21}  \\  
2   &   71.61     &   60.19      &   85.65     &   79.17     &   \textit{79.33}     &   \textit{79.33}   &  \textbf{79.58} \\ 
3   &   94.4      &   77.30        &    93.86    &   96.10     &   96.40     &  \textbf{97.45}  &   \textit{96.85} \\ 
4   &   74.25     &   49.02        &  66.77      &   81.71     &   \textit{82.62}     &   \textbf{83.36}  &   \textbf{83.36} \\ 
5   &   90.37     &   83.09        &   93.83     &   94.39     &   94.47     &   \textit{94.72}   &  \textbf{95.05} \\ 
6   &   84.92     &   50.46        &   80.43     &   \textbf{99.32}     &   \textit{98.98}     &    97.97  & \textit{98.98}  \\ 
7   &    \textbf{84.73}    &    30.93       &   72.21     &   81.58     &    82.23    &   \textit{83.44}   &  82.71 \\ 
8   &   52.01     &   50.53        &   70.96     &   70.29     &    71.66    &   \textit{71.91}   &  \textbf{72.16} \\ 
9   &   70.47     &   46.58        &   71.18     &   76.27     &    76.27    &   \textit{76.6}   &  \textbf{77.17} \\ 
10  &    92.15    &   69.73        &   96.43     &   96.49     &   96.74     &   \textit{97.08}  &  \textbf{97.83}  \\ 
11  &    \textit{89.39}    &    50.06       &   \textbf{93.73}     &   73.36     &   72.95     &   72.61  &  73.78  \\ 
12  &   60.04     &    63.44       &   \textbf{87.95}     &   \textit{75.39}     &   75.06     &   74.40   &  74.40 \\ 
13  &   86.99     &    53.35       &   84.90     &   90.66     &   \textit{93.17}     &   \textbf{95.9}    &  \textbf{95.9}  \\ 
14  &    \textit{92.99}    &    76.73       &    \textit{92.99}    &   \textbf{100.0}     &   \textbf{100.0}     &   \textbf{100.0}   & \textbf{100.0}  \\ 
15  &    \textbf{92.83}    &   54.00        &    88.30    &    86.51    &    87.3    &   90.16    & \textit{90.95} \\ 
\hline
OA  &    78.95    &   58.35        &   83.37     &    83.15    &   83.53     &   \textit{83.96}   & \textbf{84.24}  \\ 
AA  &    81.09    &   59.03        &   84.21     &   85.55     &   85.98     &   \textit{86.50}   &  \textbf{86.79} \\ 
K   &    77.32    &   55.18        &   82.07     &    81.81    &    82.22    &   \textit{82.68}   & \textbf{82.99} \\ 
\hline
\end{tabular}
\end{table}

Tables~\ref{tab: PaviaU_20}-\ref{tab: PaviaU_30} collect the comparison results on Pavia Centre and Figure~\ref{fig: PaviaU_30} exhibits the visual results. Compared with Pavia University, all seven methods achieve higher accuracy. Network found by 3D-ANAS still shows the best performance. On Pavia Centre, the improvement brought by using inference augmentation is lower than in Pavia University. We conjecture this is due to the relatively high accuracy already achieved by 3D-ANAS and the limited space for further improvement. Even so, using OV inference augmentation strategy still improves OA, AA and K by 0.07, 0.08 and 0.1 percentage points in using 20 training samples each category (Table~\ref{tab: Pavia_20}), 0.04, 0.07 and 0.06 percentage points in using 30 training samples each category (Table~\ref{tab: Pavia_30}).  In Figure~\ref{fig: Pavia_30}, it can be clearly seen that 3D-ANAS has almost no misclassified pixels in the local enlarged regions.

Tables~\ref{tab: Houston_20}-\ref{tab: Houston_30} and Figure~\ref{fig: Houston_30} illustrates the comparison results on Houston University. Compared with last two datasets, Houston University contains more spectral bands and covers more classes. Correspondingly, all methods achieve lower classification accuracy on this dataset. The classification performance of the different methods varied widely. For example, in the experiments of using 20 training samples per class, the lowest OA, AA and K are 58.32\%, 59.03\%, 55.18\%, while the highest OA, AA and K are 84.24\%, 86.79\% and 82.99\%. The gap between these two sets of results is about 30 percentage points. 3-D Auto-CNN and 3D-ANAS have much better performances compared with others. Without using any inference augmentation strategy, 3D-ANAS achieves 83.15\% OA and 81.81\% K, which are slightly lower than 3-D Auto-CNN. Yet, AA of 3D-ANAS is 85.55\%, which is significantly higher than that of 3-D Auto-CNN. When at least one inference augmentation strategy is used, 3D-ANAS slightly beats 3-D Auto-CNN in all three metrics. Overall, in the experiment setting that 20 training samples are used for each category, 3D-ANAS achieves comparable results compared with 3-D Auto-CNN. When the training samples are increased to 30 each class, 3D-ANAS reveals obvious advantages. From Figure~\ref{fig: Houston_30}, we can see that the classification map of 1-D Auto-CNN clearly shows the outline of the structure of different building. For instance, the dark red part (Commercial, Class 8) in the local enlarged region. However, at the same time, a lot of pixels are misclassified by 1-D Auto-CNN. These misclassified pixels are distributed throughout the result map and look like impulse noise, resulting in a relatively poor visual effect. On the contrary, 3-D Auto-CNN shows a very smooth result, where the outline of the structure is lost. 3D-ANAS achieves a relatively better balance between showing good visual effects and keeping outline structures. 

\begin{table}[!t]
\setlength{\abovecaptionskip}{0.cm}
\setlength{\belowcaptionskip}{-0.cm}
\renewcommand{\arraystretch}{1.3}
\caption{Comparison Experimental Results on Houston University Using 30 Training Samples Each Category}
\label{tab: Houston_30}
\centering
\begin{tabular}{C{0.8cm}|C{0.7cm}C{0.65cm}C{0.65cm}|C{0.65cm}C{0.65cm}C{0.65cm}C{0.65cm}}
\hline
Models & 3D-LWNet & 1-D Auto-CNN & 3-D Auto-CNN & 3D-ANAS & 3D-ANAS +MS & 3D-ANAS +OV & 3D-ANAS +MS+OV \\
1   &   84.81  &    72.25       &   \textbf{87.5}     &  86.16      &   86.57     &   86.57    & \textit{87.22} \\  
2   &   80.22  &   63.96        &   77.91     &  84.31      &  84.72      &   \textit{87.5}    & \textbf{87.75} \\ 
3   &   \textbf{93.45}  &   76.61        &   \textit{92.74}     &  86.06      &   85.46     &   85.46   &  84.26 \\ 
4   &   79.74  &   51.29        &   72.65     &   79.74     &   79.57     &   \textit{81.05}   &  \textbf{81.30} \\ 
5   &   90.9   &    82.25       &   96.14     &  97.11      &    \textbf{97.44}    &    \textit{97.28}   & 97.19 \\ 
6   &   81.44  &    55.32       &   84.86     &   87.46     &   \textit{88.81}     &   \textbf{89.83}   & \textit{88.81}  \\ 
7   &   \textbf{87.83}  &    34.2       &   73.03     &   77.79     &    77.79    &    \textit{81.10}   & 80.45 \\ 
8   &   63.69  &    62.11       &   76.64     &   78.75     &   78.83     &   \textit{79.16}   &  \textbf{79.32} \\ 
9   &   \textbf{77.5}   &    49.84       &    71.1    &    73.57    &     74.55   &     74.80  & \textit{75.45} \\ 
10  &    93.26 &    64.48       &   \textbf{96.67}     &   95.15     &   95.32     &   \textit{96.41}  &  95.99  \\ 
11  &    91.04 &    50.23       &   92.31     &   91.45     &   92.70     &   \textit{94.36}   & \textbf{95.77}  \\ 
12  &    79.4  &    62.47       &   \textbf{91.48}     &   87.61     &    88.11    &    89.36   & \textit{89.86} \\ 
13  &    90.41 &    50.92       &   85.8     &   \textit{91.12}     &    \textit{91.12}    &    \textbf{94.08}   & \textbf{94.08} \\ 
14  &    90.65 &    76.07       &   90.65     &   \textbf{99.75}     &   \textbf{99.75}     &   \textbf{99.75}   &  \textit{99.25} \\ 
15  &    92.47 &   62.94        &    88.85    &  93.65      &   94.28     &   \textit{95.40}   & \textbf{95.71}  \\ 
\hline
OA  &    84.17 &   60.36        &   84.45     &   86.16     &   86.51     &   \textit{87.67}   & \textbf{87.82}  \\ 
AA  &    85.12 &    61.00       &   85.22     &  87.31      &   87.67     &   \textit{88.81}   &  \textbf{88.83} \\ 
K   &    82.96 &    57.35       &   83.25     &   85.03     &   85.42     &   \textit{86.67}   &  \textbf{86.83} \\ 
\hline
\end{tabular}
\end{table}

\begin{figure*}[!h]
\setlength{\abovecaptionskip}{0.cm}
\setlength{\belowcaptionskip}{-0.cm}
\centering
\includegraphics[width=6.6in]{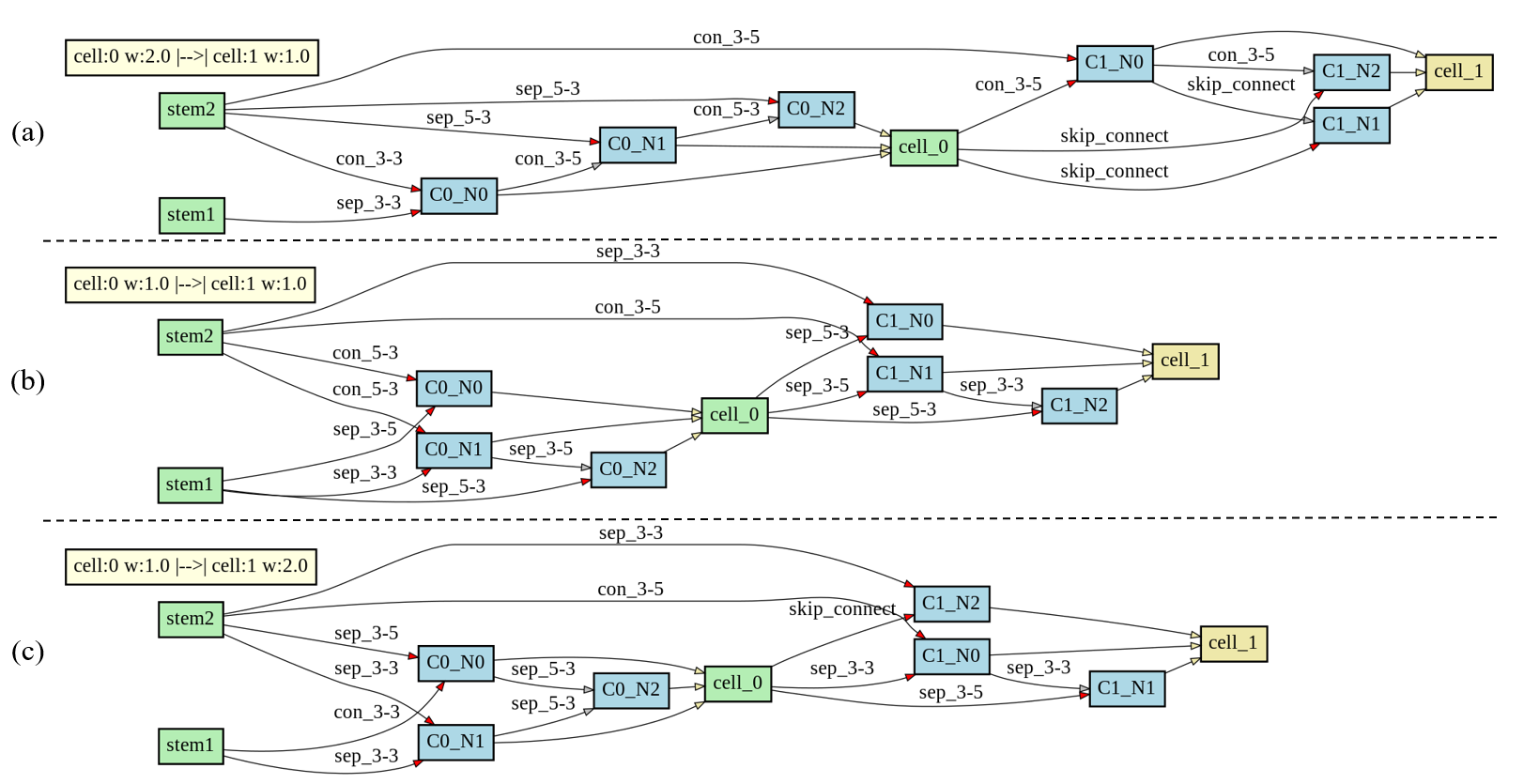}
\caption{The final architectures found for three datasets. (a) Pavia University; (b) Pavia Centre; (c) Houston University}
\label{fig: final_architectures}
\end{figure*}

\begin{figure*}[!h]
\setlength{\abovecaptionskip}{0.cm}
\setlength{\belowcaptionskip}{-0.cm}
\includegraphics[width=6.6in]{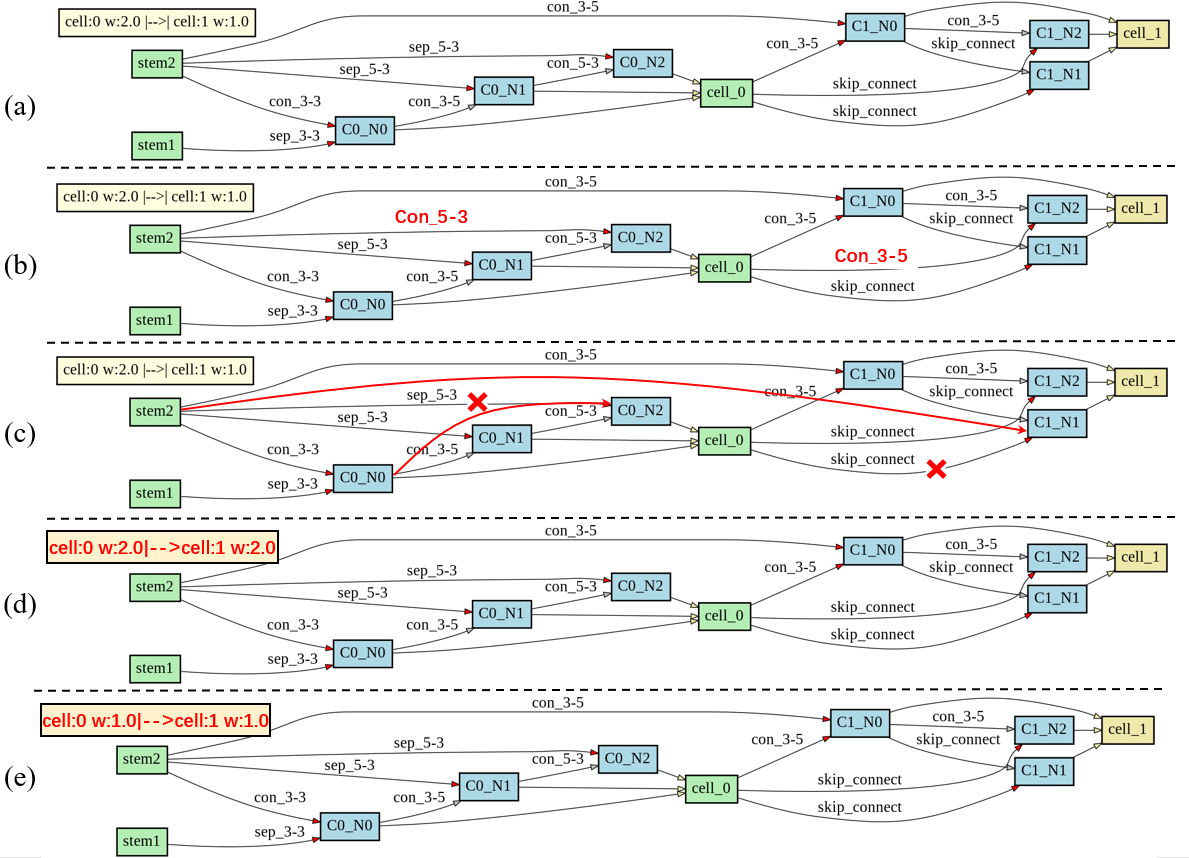}
\caption{The final architectures found for three datasets. (a) Pavia University; (b) Pavia Centre; (c) Houston University}
\label{fig: architecture_analysis}
\end{figure*}

\subsection{Architecture Analysis}
\label{sec: architecture_analysis}

In this section, we focus on analyzing the architectures found by 3D-ANAS. As the three datasets have different spectral resolution, spatial resolution and cover different objects, we seperatly perform the  architecture search for each dataset. The three found architectures are shown in Figure~\ref{fig: final_architectures}. By observing Figure~\ref{fig: final_architectures}, we note that the three architectures are different in both topological structure and selecting operations. Nevertheless, these three architectures also share some common characteristics:
\begin{enumerate}
    \item The Asymmetric convolution operation plays a major role in the final selection operations. As we introduced in Section~\ref{Sec: inner_search}, the search space we build for searching for inner topological structures contains both symmetric convolution operations which have same spatial and spectral receptive field and asymmetric convolution operations which have different spatial and spectral receptive fields. Even so, in most cases, 3D-ANAS still tend to build network which take asymmetric convolution operations as main operations and use symmetric convolution operations as complementary operations. The proportion of asymmetric convolution operations in final network designed for Paiva University is 58.33\% , for Pavia Centre is 66.67\% and for Houston University is 41.61\%. While, the proportion of symmetric convolution operations are 16.67\%, 16.67\% and 50\%. We believe this result verifies our previous analysis. The sizes of kernels along spatial dimension and spectral dimension should be different as HSIs have relatively low spatial resolution and extremely high spectral resolution. 
    \item Asymmetric convolution operations which have big spatial receptive field mainly distribute in shallow layers and asymmetric convolution operations with large spectral receptive field mainly locate in deep layers. Especially in the network designed for Pavia University, there are 7 asymmetric convolution operations. All three asymmetric convolution operations with large spatial kernel size are in the first cell and 3/4 asymmetric convolution operations with large spectral kernel size are positioned in the last cell. 
    \item 3D-ANAS prefers to set different widths for different layers. Among the three architectures, two of them have different widths for different layers.
\end{enumerate}

\subsection{Effectiveness of 3D-ANAS}

\begin{table}[]
\setlength{\abovecaptionskip}{0.cm}
\setlength{\belowcaptionskip}{-0.cm}
\renewcommand{\arraystretch}{1.3}
\caption{Effectiveness of 3D-ANAS}
\label{tab: Effectiveness_3D-ANAS}
\centering
\begin{tabular}{C{1.2cm}|C{0.45cm}C{0.8cm}C{0.8cm}C{0.9cm}C{0.8cm}C{0.8cm}}
\hline
Datasets & Acc. & 3D-ANAS & 3D-ANAS R\_op & 3D-ANAS R\_topo & 3D-ANAS R\_W$\uparrow$ & 3D-ANAS R\_W$\downarrow$ \\
\hline
\multirow{3}*{Pavia U}     & OA  & 97.81  & 97.42   & 96.84   & 97.43  & 97.18  \\
                           & AA  & 98.39  & 98.01   & 97.62   & 98.04  & 97.90  \\
                           & K   & 97.10  & 96.30   & 95.84   & 96.82  & 96.21  \\
\hline
\multirow{3}*{Pavia C}     & OA  & 99.47  & 99.13   & 98.91   & 99.03  & - \\
                           & AA  & 98.70  & 98.22   & 97.79   & 97.53  & - \\
                           & K   & 99.24  & 98.79   & 98.68   & 98.64  & - \\
\hline
\multirow{3}*{Houston U}   & OA  & 86.16  & 85.78   & 85.32   & 85.89  & 85.79  \\
                           & AA  & 87.31  & 86.96   & 86.57   & 87.01  & 86.93  \\
                           & K   & 85.03  & 84.63   & 84.31   & 84.77  & 84.65  \\
\hline
\end{tabular}
\end{table}

As shown in Figure~\ref{fig: final_architectures}, the networks found by our proposed 3D-ANAS have many fragmented branches. According to study results in~\cite{ma2018shufflenet}, these fragmentation structures are beneficial for improving classification accuracy. To verify whether 3D-ANAS improves the accuracy by really searching for efficient architectures or by simply integrating various branches and convolution operations, we manually modify the architecture automatically designed by our proposed 3D-ANAS in three different ways. Taking the architecture designed for Pavia University as an example, the three cases are shown in Figure~\ref{fig: architecture_analysis}, where the manually modified parts are marked in red. Specifically, the three modifications are: 1) randomly replacing two 3D-ANAS selected operation with others (termed as R\_op). For instance, replacing sep\_5-3 with Con\_5-3; 2) changing the connection relationships between the nodes inside each cell (termed as R\_topo); 3) increasing the widths (R\_W$\uparrow$)， all widths of cells are set to ${\gamma}^{1}W$.  Decreasing widths (R\_W$\downarrow$), all widths of cell are set to ${\gamma}^{0}W$. 

Following the three modification methods, we modify the found architectures several times. Each time different operations and connections in different nodes are modified. According to experimental results, we find that the modified architectures generally yield lower performance. The results are listed in Table~\ref{tab: Effectiveness_3D-ANAS}, according to which we can draw the following conclusions:

\begin{enumerate}
    \item 3D-ANAS is able to select a proper operation for each path. Comparing the results in the fourth column with those in the third column, we can see that replacing the automatically select operations with others decreases accuracy on all three datasets. 
    \item 3D-ANAS has the ability to design efficient topological structures. For instance, on Pavia University, the modification in topological structure drops OA, AA and K to 96.84\%, 97.62\% and 95.84\%, respectively.  
    \item Compared with randomly replacing selected operations, changing in network architecture impacts the efficiency of network more seriously. 
    \item Searching widths for each cell is beneficial for improving classification accuracy. Using either R\_W$\uparrow$ or R\_W$\downarrow$ shows a slightly negative on the performance. With manually designed widths, the performance is still decent. Besides, searching widths can further improve the performance.   
\end{enumerate}

Overall, the experimental results demonstrate that 3D-ANAS does find an efficient topological structure and selects proper operation, rather than simply integrating a network with many branches and various convolution operations. The fact that a slight perturbation to the found architecture deteriorates the accuracy indicates that the found architecture is indeed a local optimum in the architecture search space.

\subsection{Effectiveness of 3D Asymmetric Search Space}

\begin{table}[]
\setlength{\abovecaptionskip}{0.cm}
\setlength{\belowcaptionskip}{-0.cm}
\renewcommand{\arraystretch}{1.3}
\caption{Comparison experiments between different search spaces}
\label{tab: comparision_search_space}
\centering
\begin{tabular}{C{0.9cm}C{1.7cm}|C{0.4cm}C{0.7cm}C{0.65cm}C{0.65cm}C{0.65cm}}
\hline
Datasets & Search spaces & de. &  asym. &  OA   & AA   &   K \\
\hline
\multirow{4}*{Pavia U.}   & 3D-sym-ud      & \ding{55}  &  \ding{55}  & 96.91  & 97.24  & 95.91  \\
                          & 3D-sym-d    & \ding{51}  &  \ding{55}     & 97.17  & 97.71  & 96.26  \\
                          & 3D-asym-ud  & \ding{55}  &  \ding{51}     & 97.45  & 98.11  & 97.03  \\
                          & 3D-asym-d   & \ding{51}  &  \ding{51}     & 97.81  & 98.39  & 97.10  \\
\hline
\multirow{4}*{Pavia C.}   & 3D-sym-ud      & \ding{55}  &  \ding{55} & 99.02 & 98.19 & 98.61 \\
                          & 3D-sym-d    & \ding{51}  &  \ding{55}  &   99.13 & 98.09 & 98.76 \\
                          & 3D-asym-ud  & \ding{55}  &  \ding{51}  &  99.31 & 98.55  & 99.03 \\
                          & 3D-asym-d   & \ding{51}  &  \ding{51}  &  99.47 & 98.70  & 99.24 \\
\hline
\multirow{4}*{Houston U.} & 3D-sym-ud   & \ding{55}  &  \ding{55}  &  85.21 & 86.32  &  84.19 \\
                          & 3D-sym-d    & \ding{51}  &  \ding{55}  &  85.63 & 86.91  & 84.59 \\
                          & 3D-asym-ud  & \ding{55}  &  \ding{51}  &  85.91 & 87.21  & 84.76 \\
                          & 3D-asym-d   & \ding{51}  &  \ding{51}  &  86.16 & 87.31  & 85.03 \\
\hline
\end{tabular}
\end{table}

\begin{figure*}[!t]
\setlength{\abovecaptionskip}{0.cm}
\setlength{\belowcaptionskip}{-0.cm}
\centering
\includegraphics[width=7in]{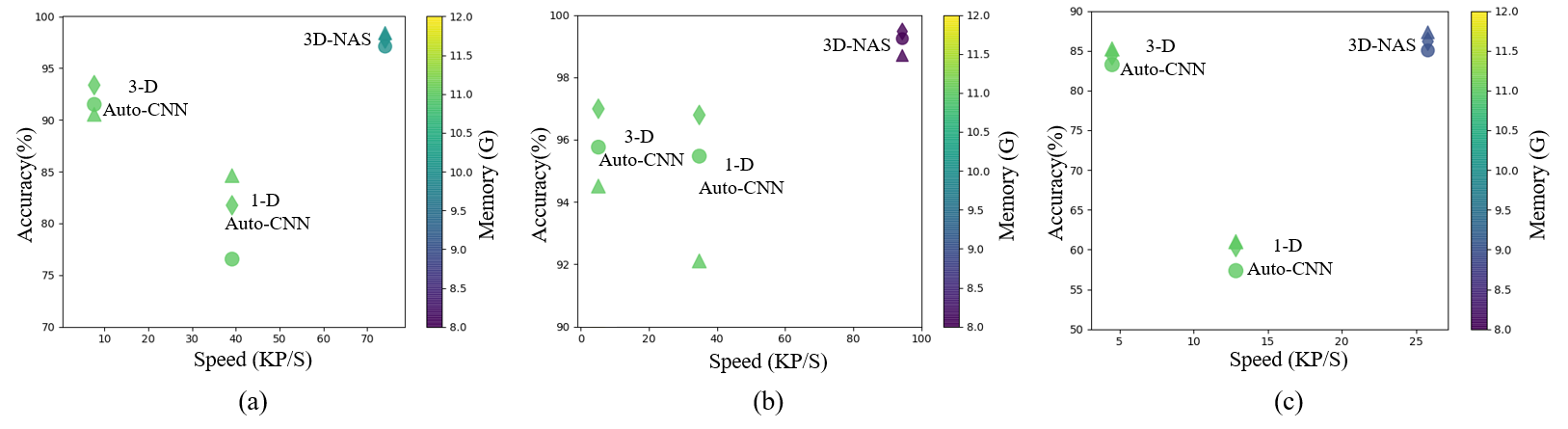}
\caption{The final architectures found for three datasets. (a) Pavia University; (b) Pavia Centre; (c) Houston University. OA, AA and K are marked with rhombic, triangular and circular icons. }
\label{fig: inference_speed}
\end{figure*}

By in-depth analyzing HSIs, we find that spectral resolution and spatial resolution are not consistent in HSIs. The spatial resolution is relatively low while the spectral resolution is extremely high. Based on this discovery, we conjecture that spatial dimension and spectral dimension should be processed differently. So we build a 3D asymmetric decomposition search space, which contains three kinds of convolution operations: 1) asymmetric convolution operation where kernel size along spatial dimension is larger than that along spectral dimension; 2) asymmetric convolution operation which is the opposite of 1). 3) decomposed convolution operation where, the kernel size along spatial and spectral dimension are the same. In section~\ref{sec: architecture_analysis}, the searching results have demonstrated that 3D-ANAS does prefer to select 3D asymmetric convolution operations, which partly supports our conjecture. In this section, to further evaluate the effectiveness of 3D asymmetric decomposition search space, we build three other different search spaces, including 1) 3D symmetrically undecomposed search space (3D-asym); 2) 3D symmetrically decomposed search space (3D-sym-d); 3) 3D asymmetrically undecomposed search space (3D-asym-ud). The specific components of these three search spaces are listed as follows:

\vspace{3pt}
\noindent{\bf 3D symmetrically undecomposed search space}
\begin{itemize}
    \item Con\_3: LReLU-$\rm Conv(3\times3\times3)$-BN;
    \item Con\_5: LReLU-$\rm Conv(5\times5\times5)$-BN;
    \item Con\_7: LReLU-$\rm Conv(7\times7\times7)$-BN;
    \item Sep\_3: LReLU-$\rm Sep(3\times3\times3)$-BN;
    \item Sep\_5: LReLU-$\rm Sep(5\times5\times5)$-BN;
    \item Sep\_7: LReLU-$\rm Sep(7\times7\times7)$-BN;
    \item skip\_connection: $f(x)=x$;
    \item discarding: $f(x)=0$.
\end{itemize}

\vspace{3pt}
\noindent{\bf 3D symmetrically decomposed search space}
\begin{itemize}
    \item Dcon\_3-3: LReLU-$\rm Conv(1\times3\times3)$-$\rm Conv(3\times1\times1)$-BN;
    \item Dcon\_5-5: LReLU-$\rm Conv(1\times5\times5)$-$\rm Conv(5\times1\times1)$-BN;
    \item Dcon\_7-7: LReLU-$\rm Conv(1\times7\times7)$-$\rm Conv(7\times1\times1)$-BN;
    \item Dsep\_3-3: LReLU-$\rm Sep(1\times3\times3)$-$\rm Sep(3\times1\times1)$-BN;
    \item Dsep\_5-5: LReLU-$\rm Sep(1\times5\times5)$-$\rm Sep(5\times1\times1)$-BN;
    \item Dsep\_7-7: LReLU-$\rm Sep(1\times7\times7)$-$\rm Sep(7\times1\times1)$-BN;
    \item skip\_connection: $f(x)=x$;
    \item discarding: $f(x)=0$.
\end{itemize}

\vspace{3pt}
\noindent{\bf 3D asymmetrically undecomposed search space}
\begin{itemize}
    \item Udcon\_3-3: LReLU-$\rm Conv(1\times3\times3)$-$\rm Conv(3\times1\times1)$-BN;
    \item Udcon\_5-5: LReLU-$\rm Conv(1\times5\times5)$-$\rm Conv(5\times1\times1)$-BN;
    \item Udcon\_7-7: LReLU-$\rm Conv(1\times7\times7)$-$\rm Conv(7\times1\times1)$-BN;
    \item Udsep\_3-3: LReLU-$\rm Sep(1\times3\times3)$-$\rm Sep(3\times1\times1)$-BN;
    \item Udsep\_5-5: LReLU-$\rm Sep(1\times5\times5)$-$\rm Sep(5\times1\times1)$-BN;
    \item Udsep\_7-7: LReLU-$\rm Sep(1\times7\times7)$-$\rm Sep(7\times1\times1)$-BN;
    \item skip\_connection: $f(x)=x$;
    \item discarding: $f(x)=0$.
\end{itemize}

In 3D convolution search space, the 3D convolution operations are common 3D convolutions, which contains two kinds of convolutions, conventional 3D convolution and separable 3D convolution, each of which contains three specific setting corresponding to different kernel sizes. In the 3D symmetric search space, each conventional 3D convolution is decomposed into a sequence of two convolutions regarding to spatial convolution and spectral convolution. 

The comparison results between these three search spaces are listed in Table~\ref{tab: comparision_search_space}. The architectures found by using the 3D convolution search space have lower accuracy compared with the architectures designed by using other two search space. Architectures found by using symmetric search space have worse performance compared with architectures obtained by adopting asymmetric search space. For Pavia University, decomposing 3D convolution in 3D convolution search space improve overall accuracy by about 0.3 percentage points and introducing and asymmetric structure improves the the overall accuracy by about 0.6 percentage points. From 3D-sym-ud to 3D-asym-d, OA, AA and K are increased by 0.9, 1.15 and 1.21 percentage points, respectively. 

\subsection{Inference Speed}

Previous deep learning based HSI classification methods usually employ a patch-to-pixel classification framework, which is relatively inefficient in computation overlapping regions of adjacent patches need to be repeatedly computed. To overcome this issue, in our 3D-ANAS, we employ a pixel-to-pixel classification framework instead. In this section, we compare the inference speed of these two classification frameworks and show the comparison results in Figure~\ref{fig: inference_speed}.

In the comparison experiments of using 30 training sample each class, our proposed 3D-ANAS achieves the best performances on all three datasets, while having the fast inference speed. Especially on Paiva Centre (Figure~\ref{fig: inference_speed} (b)), 3D-ANAS has 94.37 KP/S (thousand pixel per second) inference speed, which is about 18.15$\times$ that of 3-D Auto-CNN and 2.71$\times$ that of 1-D Auto-CNN. it is worth noting that,  1-D Auto-CNN based HSI classification is also a pixel-to-pixel classification, without duplicate processing pixels. However, the inference speed of 1-D Auto-CNN is still lower than our 3D-ANAS. We conjecture this is caused by the fact that the operate mode in 3D-ANAS is more aligned with Pytorch framework. 

\section{Conclusion}

Most deep learning HSI classification methods are based on manually designed networks, in which a lot of human efforts are spent on designing state-of-the-art neural network architectures. To overcome this inhibitor, we have proposed 3D-ANAS, an asymmetric neural architecture search algorithm for HSI classification. The proposed 3D-ANAS has a hierarchical search space, including an inner search space and an outer search space. In the inner search space, a 3D asymmetric search space is build according to characteristics of HSIs for searching for architectures. In the outer search space, three candidate paths with different widths are provided for searching for width for each layer. In addition, to improve the inference speed and overcome the problems of repetitive operations and scale sensitivity in the patch-to-pixel classification framework, 3D-ANAS employs a new classificaion framework, which is a pixel-to-pixel classification framework. 

We conduct comparison experiments on three challenging public HSI datasets and compare our proposed 3D-ANAS with other deep learning based HSI classification methods, including both manually designed CNN based method (3D-LWNet) and automatic design CNN based methods (1-D Auto-CNN and 3-D Auto-CNN). The results of comparison experiments have demonstrated our proposed 3D-ANAS achieve competitive performance with others. In addition, we have organized abundant ablation experiments to verify the effectiveness of our proposed 3D asymmetric search space, 3D-ANAS and pixel-to-pixel classification framework. Results of the ablation study show that our 3D-ANAS does find a local optimum architecture in the architecture search space and pixel-to-pixel classification is beneficial for improving inference speed. Compared with patch-to-pixel classification framework, pixel-to-pixel classification framework improves inference speed up to 18$\times$. 

In our future work, we will focus on proposing more efficient neural architecture search algorithms to automatically design efficient architectures for HSI classification. In addition, we will pay more attention to study the relationship between the found architectures and characteristics of HSIs. 


%

\appendices



\ifCLASSOPTIONcaptionsoff
  \newpage
\fi



%


\bibliographystyle{IEEEtran}
\bibliography{IEEEabrv,reference}
%







\end{document}